\documentclass{article}

\usepackage{microtype}
\usepackage{graphicx}
\usepackage{subcaption}

\usepackage{booktabs} 

\usepackage[preprint]{icml2026}

\usepackage{amsmath}

\usepackage{amssymb}
\usepackage{mathtools}
\usepackage{amsthm}

\usepackage[textsize=tiny]{todonotes}

\usepackage[utf8]{inputenc}
\usepackage[T1]{fontenc}
\usepackage{url}
\providecommand{\Url}[1]{\url{#1}}
\providecommand{\url}[1]{\texttt{#1}}
\usepackage{booktabs}       %
\usepackage{amsfonts}       %
\usepackage{microtype}      %
\usepackage{xcolor}         %
\usepackage{amsmath}      %
\usepackage{algorithm}

\usepackage{algpseudocode}
\usepackage{amssymb}      %
\usepackage{natbib}
\usepackage{adjustbox}
\usepackage{array}
\usepackage{caption}
\usepackage{makecell}
\usepackage{epigraph}
\usepackage{float}
\usepackage{enumitem}
\usepackage{listings}
\usepackage{wrapfig}
\usepackage{xspace}
\usepackage{subcaption}
\usepackage{multirow,wasysym}
\usepackage{siunitx}

\definecolor{linkcolor}{RGB}{74, 102, 146}
\usepackage[
colorlinks=true,allcolors=linkcolor,pageanchor=true,
plainpages=false,pdfpagelabels,bookmarks,breaklinks,bookmarksnumbered,
backref=page
]{hyperref}
\usepackage[capitalize,noabbrev]{cleveref}

\definecolor{lightpurple}{RGB}{168, 141, 201}

\definecolor{darkgreen}{RGB}{0,128,0}
\newcommand{\vect}[1]{\boldsymbol{#1}}
\newcommand{\mat}[1]{\mathbf{#1}}
\newcommand{\R}{\mathbb{R}}

\newcommand{\q}{\vect{q}}
\newcommand{\kvec}{\vect{k}}
\newcommand{\vvec}{\vect{v}}
\newcommand{\x}{\vect{x}}

\newcommand{\h}{\vect{h}}
\newcommand{\ovec}{\vect{o}}

\newcommand{\K}{\mat{K}}
\newcommand{\V}{\mat{V}}
\newcommand{\Q}{\mat{Q}}
\newcommand{\M}{\mat{M}}
\newcommand{\Ob}{\mat{O}}

\newcommand{\WQ}{\mat{W}_Q}
\newcommand{\WK}{\mat{W}_K}
\newcommand{\WV}{\mat{W}_V}

\newcommand{\alphab}{\boldsymbol{\alpha}}
\newcommand{\thetab}{\boldsymbol{\theta}}
\DeclareMathOperator{\softmax}{softmax}

\icmltitlerunning{Sample-Efficient In-Context Adaptation via Sleep}

\begin{document}

\twocolumn[
  \icmltitle{Sample-Efficient In-Context Adaptation via Sleep}



  \icmlsetsymbol{equal}{*}
 \icmlsetsymbol{internship}{\textdagger}
  \begin{icmlauthorlist}
    \icmlauthor{Sangyun Lee}{yyy,internship}
    \icmlauthor{Zhang Enyan}{zzz}
    \icmlauthor{Michael Xu}{xxx}
    \icmlauthor{Hao Tang}{xxx}
    \icmlauthor{Zinan Lin}{xxx}
    \icmlauthor{Weijia Xu}{xxx}
    \icmlauthor{Chris Quirk}{xxx}
    \icmlauthor{Jianfeng Gao}{xxx}
    \icmlauthor{Giulia Fanti}{yyy}
    \icmlauthor{Xiaodong Liu}{xxx}
  \end{icmlauthorlist}

  \icmlaffiliation{yyy}{Carnegie Mellon University}
\icmlaffiliation{xxx}{Microsoft Research}
\icmlaffiliation{zzz}{Yale University}

  \icmlcorrespondingauthor{Sangyun Lee}{sangyunl@andrew.cmu.edu}

  \icmlkeywords{Machine Learning, ICML}

  \vskip 0.3in
]



\printAffiliationsAndNotice{\textsuperscript{\textdagger}\,The initial phase of this work was
  conducted during SL's summer internship at Microsoft Research.\par\rightskip=0pt plus 2em}  

\begin{abstract}
The ability to learn and generalize from limited experience, or \textit{sample efficiency}, is a cornerstone of intelligence.
How can we build language models that turn a few user interactions, a few snippets of unfamiliar code, or a handful of experimental results provided in context into knowledge that generalizes?
Prior work has found that \textit{sleep}, a process where the model performs $N$ recurrent forward passes to refine its representations of a chunk of context, can improve long-context reasoning (specifically, within SSM-attention hybrid models). We study whether this method also improves \emph{in-context adaptation} on general, non-synthetic tasks for  transformer language models.
We pre-train language models up to 1.2B-parameters on up to 100B tokens with sleep mechanisms. 
Notably, longer sleep (i.e., using a larger number of recurrent forward passes $N$) enables the model to learn faster in-context on multiple few-shot algorithmic, question-answering tasks, while also achieving lower next-token prediction loss on out-of-distribution evaluations.
On held-out code, models with $N=2$ and $N=4$ match the
no-sleep baseline's loss using $2.10\times$ and $2.39\times$
fewer tokens in-context, respectively.
More broadly, we view our work as a step toward building models that have better tradeoffs between in-context adaptation and decoding latency.

\end{abstract}

\section{Introduction}

\begin{figure}
    \centering
    \includegraphics[width=1\linewidth]{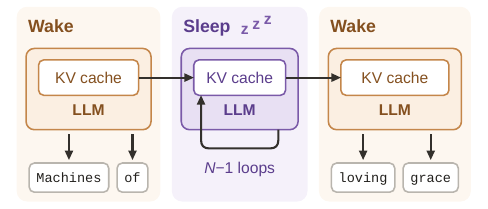}
    \caption{During the wake phase, our model decodes tokens at a normal speed without using any depth recurrence. Once the sleep boundary is reached (i.e., once the sleep window is full), the model performs additional recurrent passes over the window tokens to refine the KV cache. It then resumes decoding at a normal speed. This is done both during training and at inference time.
    }
    \label{fig:decode}
\end{figure}

\begin{figure*}[t]
    \centering
    \includegraphics[width=1\linewidth]{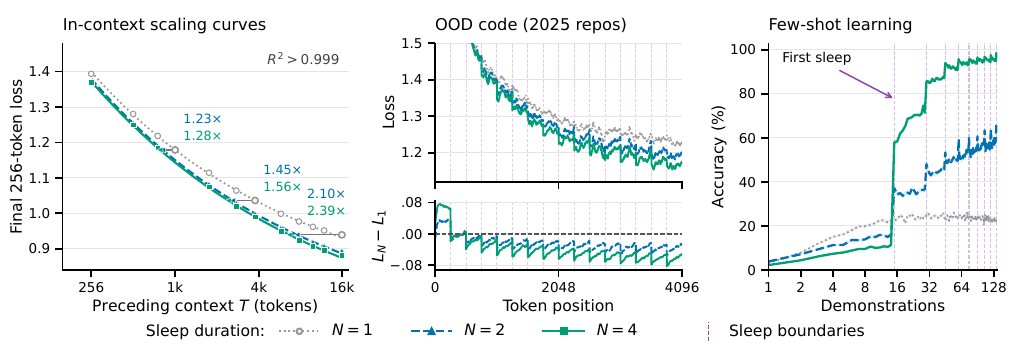}
    \vspace{-5mm}
    \caption{\textbf{Longer sleep (i.e., a larger $N$) leads to better in-context adaptation.} Left: Mean next-token prediction loss on the final 256 tokens from an out-of-distribution (OOD) code dataset with varying prefix  length. Models with sleep reach losses of models with no sleep at faster rates, and the sample efficiency gain increases as context grows.    Middle: Loss per token position on the OOD code dataset. Sleep models have higher loss than the no-sleep baseline in the first window, and loss improves steeply at sleep boundaries (shown as vertical dashed lines). Right: On a few-shot learning task (selective recall), models with longer sleep duration are significantly better than those with less or no sleep.
    }
    \label{fig:main}
    \vspace{-5mm}
\end{figure*}

The ability to learn and generalize from limited experience, or \textit{sample efficiency}, is a cornerstone of intelligence~\citep{chollet2019measure}.
Human learning is extremely sample efficient: a generous upper estimate of linguistic exposure over ten years is only 1.4 billion tokens~\citep{brysbaert2016many}.
Yet within that time, a teenager can become a professional mathematician, chess player, rocket scientist, or a programmer, among many other professions.
Can language models learn from new experience as efficiently as humans?
In contrast with pre-training, where language models are many orders of magnitude less sample-efficient than humans, language models learn with much better efficiency via \emph{in-context adaptation}: at deploy-time, language models can adapt their behavior based on its preceding input (also known as ``context''), such as material published after their training-data cutoff, user preferences, and environment feedback.
Yet, a substantial gap still remains between their sample efficiency and that of humans~\citep{liu2026agents}.

\textbf{Model scaling improves sample efficiency at a high cost.}
One way to improve the sample efficiency of in-context adaptation is to increase model size.
\citet{kaplan2020scaling} show that per-token loss decreases following a predictable power law, and larger models achieve a faster decrease in loss as context accumulates.\footnote{Note that this is a separate result from more widely known scaling law results from the same paper, which describes average loss as a function of model/dataset size.} This observation is later validated at scale by \citet{brown2020language} on several in-context adaptation benchmarks.

However, larger models increase both training and inference costs.
Inference cost is particularly limiting because it recurs with every use,
whereas training cost can be amortized over the model's deployment lifetime.
For instance, doubling the model depth doubles KV cache size and weight size, and approximately doubles per-token decoding latency. 
Demand for inference is growing rapidly, with inference-time compute soon projected to dominate the compute market; for instance, OpenAI reported that they process more than 20 trillion tokens per day~\citep{openai2026nextphase} in March 2026, which is about 2.5 times its reported volume six months earlier \cite{openai2025devday}.\footnote{\citet{openai2026nextphase} wrote ``Our APIs now process more than 15 billion tokens per minute.'' \citet{openai2025devday} wrote ``6B tokens per minute on the API platform''.}
This cost also matters during reinforcement-learning post-training,
where rollout generation accounts for much of the computation
.
Larger models also have a larger memory footprint, limiting the hardware on which they can run.
For instance, Kimi K3~\citep{team2026kimi} has 2.8 trillion parameters, and its released weights do not fit in the GPU memory of a node with eight NVIDIA B200s without additional quantization.

\textbf{Sleep as a mechanism for in-context adaptation.}
We introduce a transformer-based  model architecture that learns more effectively from context without increasing the model size. As shown in \Cref{fig:decode}, the model periodically goes to \emph{sleep}~\citep{lee2026language}, using recurrent forward passes over a chunk of context to iteratively refine its key-value cache, thereby achieving better in-context adaptation. 
{While \citet{lee2026language} have shown that sleep improves long-range reasoning on synthetic datasets, (a) they do not evaluate in-context adaptation on general, non-synthetic datasets, focusing instead on reasoning about evicted context, and (b) their experiments are on hybrid state-space model (SSM)-attention models. Our paper adapts the sleep idea to transformer architectures and, more importantly, demonstrates its efficacy for in-context adaptation on general tasks.}

\textbf{Result: In-context sample efficiency.}
\Cref{fig:main} demonstrates  in-context adaptation performance as a model processes more context, tested on a 1.2B parameter model trained on 32B tokens on DCLM~\citep{li2024datacomp} with different sleep durations $N$. We observe that across tasks of different nature and complexity, from algorithmic in-context adaptation tasks to next-token prediction on unseen code, \textbf{models that \emph{sleep} exhibit markedly better in-context sample efficiency than models that do not, and this gap increases with a longer sleep duration $N$}. In the leftmost panel, we see that at a 16k context token cap, a model with $N=2$ and $N=4$ sleep duration achieves equal next-token prediction loss with 16k prefix tokens with $2.10\times$ and $2.39\times$ less context, respectively.
As shown in the middle and the rightmost panels, sleep models have higher loss/lower accuracy than the no-sleep baseline in the first window, when little context is available and predictions rely more on knowledge stored in the weights.
Their advantage grows as context accumulates over successive sleep phases, suggesting that using sleep during the training and inference yields a model that adapts better to new context and relies less on memorized knowledge.

\textbf{Result: Performance-efficiency Pareto frontier.}
Compared to a no-sleep variant, the model adds less than $2\%$ to the parameter count, with no increase in KV-cache size. Additionally,
since the recurrent passes are performed offline during sleep, it incurs no additional wake-time latency overhead.
Overall, including sleep-time overhead, end-to-end decoding latency increases by less than $3\%$ with one sleep iteration and $8\%$ for three sleep iteration with 128k prefix for 1.2B models, allowing us to achieve Pareto-frontier compared to other looped transformer baselines (\Cref{sec:ablation-baseline}). In aggregate, our results suggest that \emph{sleep} may be a promising direction for improving in-context adaptation abilities of transformers, while incurring limited additional overhead in terms of inference latency.

\section{Background and related work}
\subsection{Attention}
Attention~\citep{vaswani2017attention} is a sequence-mixing operation where token representations are mixed according to query-key similarity.
For the token representation $\h_t$ at timestep $t$, define
\begin{align}
\q_t &= \WQ \h_t, &
\kvec_t &= \WK \h_t, &
\vvec_t &= \WV \h_t,
\end{align}
where $\q_t,\kvec_t,\vvec_t \in\R^d$ 
are column vectors, and $\WQ,\WK,\WV$ are learned projection matrices.
For generating an output vector $\ovec_t$ for the $t$-th token,
self-attention stores all previous keys $\kvec_t$ and values $\vvec_t$ in $\K_{1:t}=[\kvec_1,\ldots,\kvec_t]^\top\in\R^{t\times d}$ and $\V_{1:t}=[\vvec_1,\ldots,\vvec_t]^\top\in\R^{t\times d}$, then computes
\begin{align}
\ovec_t
&=
\V_{1:t}^\top
\softmax\!\left(\frac{\K_{1:t} \q_t}{\sqrt d}\right).
\end{align}
Here, $\K_{1:t}$ and $\V_{1:t}$ are called the key-value (KV) cache.

For parallel computation, let $\Q_{1:t}=[\q_1, \ldots, \q_t]^\top$ and $\Ob_{1:t}=[\ovec_1, \ldots, \ovec_t]^\top$ and compute
\begin{align}
    \Ob_{1:t}=
\softmax\!\left(
\frac{\Q_{1:t}\K_{1:t}^\top}{\sqrt d}+\M
\right)\V_{1:t},
\end{align}
where $\softmax$ is applied row-wise and $\M$ is an additive causal mask.

\subsection{Sample efficiency}
Prior work attempts to improve the data efficiency of language-model pre-training by increasing training epochs, model scaling, and ensembles of independently trained models~\citep{lovelace2026prescriptive,mandal2026q0,
muennighoff2023scaling,kim2026pre}.
We instead study sample efficiency during in-context adaptation by scaling computational depth.

\subsection{In-context adaptation and computational depth}
\citet{kaplan2020scaling} analyze the next-token prediction loss at each token position and show that bigger models have steeper in-context adaptation curves.
Although we have yet to fully understand why this happens (rather than, for instance, having a curve with the same slope but only a different offset),  existing evidence suggests that some in-context adaptation algorithms benefit from sufficient computational depth.
For example, the standard induction head circuit~\cite{olsson2022context}, which performs associative in-context recall, can emerge only when two or more attention operations are composed.
$k$-hop induction appears to require at least $\log k$ compositions of attention~\cite{sanford2024transformers}.
Function vectors \cite{hendel2023context,todd2024function}, a more complex form of in-context adaptation, can emerge when the model has enough depth to extract and apply the task vector.
\citet{von2023transformers} show that transformers with multiple layers can emulate multi-step gradient descent in linear regression tasks.
These observations motivate the hypothesis that increasing computational depth is key for sophisticated in-context adaptation algorithms to emerge.
However, token generation latency grows linearly with the number of layers, limiting  how far the model depth can be increased.

\begin{figure*}
    \centering
    \includegraphics[width=\linewidth]{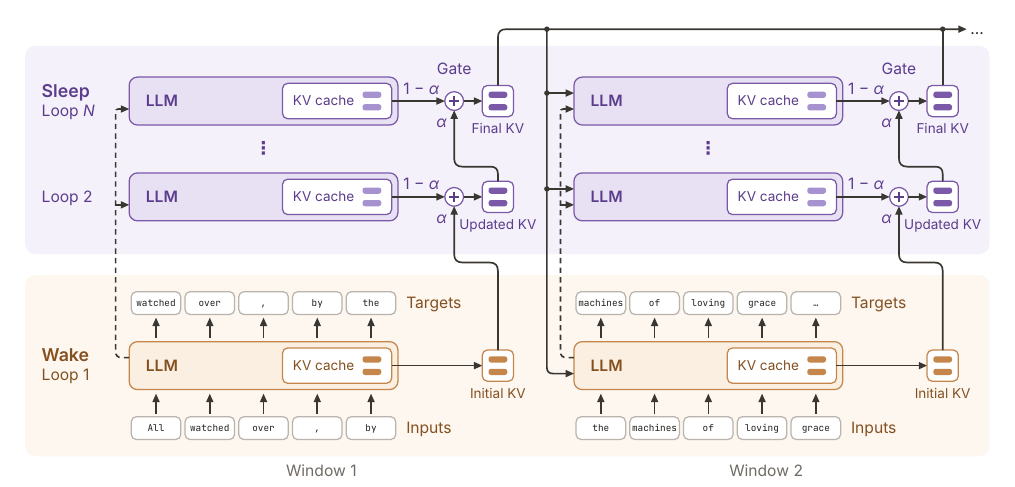}
    \caption{Training transformers with sleep. The first loop of the language model is used to produce the next-token predictions. Additional loops (``sleep''), performed only at chunk boundaries, are used refine the completed window's KV cache.}
    \label{fig:arch}
\end{figure*}

\subsection{Depth-recurrent models}
Depth-recurrence (or ``looping'') is one major pathway to increasing depth without increasing the number of parameters \citep{dehghani2018universal,graves2016adaptive,elbayad2019depth,schwarzschild2021can,bansal2022end}.
Recent work applies depth recurrence to language models through
pretraining~\citep{geiping2026scaling,zhu2025scaling} or
retrofitting pretrained models~\citep{mcleish2025teaching}.
Other studies examine design choices that improve optimization
and performance~\citep{prairie2026parcae,schwethelm2026much,
huang2026towards,saunshi2025reasoning,wang2026smelt}.
Although depth recurrence reduces parameter memory, its decoding latency and KV cache memory grow linearly with the number of loops.
Inspired by methods that share KV cache across layers in non-recurrent transformers~\citep{brandon2024reducing,sun2024you,wu2024layer}, prior work has proposed sharing the KV cache across loops either in a training-free manner~\citep{zhu2025scaling,geiping2026scaling} or with training~\citep{wu2025parallel,vendrell2026memory,bae2026mixture}.
Among these approaches, \citet{vendrell2026memory} is closest to ours. We discuss their approach in \Cref{sec:method} and compare it with ours in \Cref{sec:additional-analyses}.

\subsection{Sleep}
Several machine learning methods~\citep{hafner2019dream,sutton1991dyna,ha2018world,mnih2015human,hinton1995wake,carreira2005contrastive,lin2025sleep,behrouzlanguage} use offline processing phases inspired by the role of biological sleep in animal learning~\citep{rasch2013sleep,mcclelland1995there}.
Recently, \citet{lee2026language} proposed a sleep-like depth-recurrence mechanism where the model converts recent context into fast weights before clearing its key-value cache.
Once the context size reaches the sleep window size $W$, the model goes to sleep, during which it performs $N-1$ additional depth-recurrent steps to iteratively refine fast weights, providing enough computational depth for learning.
After sleep, the model evicts context tokens and resumes decoding at normal latency, incurring no additional cost between sleeps.
On synthetic tasks, they show that increasing sleep duration $N$ improves the model's ability to retrieve and reason about evicted context.
However, \citet{lee2026language} conducted experiments on synthetic datasets explicitly designed to evaluate reasoning over evicted context, and it is unclear whether this same mechanism can (a) improve \emph{in-context adaptation} (rather than just reasoning) of language models trained on web text, and (b) do so even \emph{without} context eviction.
In this paper, we study whether this mechanism can be scaled to transformer pre-training to improve general in-context sample-efficiency.

Motivated by the memory consolidation process in the brain, \citet{oomerjee2026bottlenecked} add a lightweight adapter network to a frozen language model to refine its KV cache after the frozen model predicts each chunk of tokens.
The adapter is trained to minimize the next-token prediction loss on the following chunk, without backpropagating losses from later chunks through the current refinement, as they use truncated backpropagation through time.
They train the adapter on supervised reasoning traces and observe that it improves performance on downstream reasoning tasks.
Sleep also refines the KV cache, but instead of introducing a lightweight adapter, we use the main network itself to iteratively refine the cache through depth recurrence, increasing computational depth and the expressivity of the update.
We also do not truncate gradients across windows.
Our focus is the sample efficiency of in-context adaptation, and whether increasing $N$ can yield further sample efficiency gains while accounting for inference cost.

\subsection{Amortizing training compute during inference}

When inference demand is large, spending more compute during training
can be worthwhile because that cost is amortized over the model's
deployment lifetime.
One approach is to train smaller models well beyond the
Chinchilla-optimal token budget~\citep{sardana2023beyond}.
For example, Qwen2-0.5B is pretrained on 12T
tokens~\citep{Yang2024Qwen2TR}, approximately 1,200 times the Chinchilla-optimal token budget.
However, the supply of useful training data limits how far
overtraining can extend~\citep{villalobos2022will,kim2026pre,sutskever2024whatadecade}.
This motivates methods that spend more computation per training
example while preserving inference cost.

Recent temporal-recurrence architectures illustrate this
approach~\citep{cai2026t,wang2026full,huang2026latent}.
They use additional parallel refinement steps during training
to approximate recurrent computations across token positions.
During autoregressive decoding, the preceding token's recurrent
state is already available, so each new token requires only one
forward pass.
These methods therefore spend additional training compute while
retaining approximately the decoding cost of a standard Transformer.

Sleep follows a similar philosophy: it spends approximately $N$
times the training compute of the no-sleep baseline to learn more effective in-context adaptation but concentrates
additional compute into periodic offline phases, leaving wake-time decoding latency unchanged.
In \Cref{sec:inference}, we show that the resulting end-to-end
decoding overhead is marginal in our benchmark.

\begin{figure*}[t]
\centering
\begingroup
\definecolor{exampleWake}{RGB}{255,245,231}
\definecolor{exampleSleep}{RGB}{244,239,251}
\setlength{\fboxsep}{6pt}
\setlength{\fboxrule}{0.4pt}
\fcolorbox{black!25}{black!2}{%
\begin{minipage}{0.96\textwidth}
\small
\setlength{\parindent}{0pt}
\setlength{\parskip}{3pt}
\textbf{Example: one-layer transformer with sleep}

Consider processing the first 6 tokens $x_1,\ldots,x_6$ in a sequence with the sleep window size $W=3$.
Let our model be a simplified one-layer, single-head  transformer without MLPs, normalization, or positional encodings.
$\h_t^l$ and $\ovec_t^l$ are its attention input and output at each loop $l$.
First, we start by obtaining token embeddings via $\h_{t}^1 = \operatorname{Embed}(\x_{t})$ for all $t$ in order to obtain the first loop input. Initialize $\mathcal L$ with zero. Then, we proceed to step 1 and then to step 2. During training, after the forward pass finishes, we backpropagate through the entire computation {(including recursions)} and update the model parameters.

\medskip
\begin{minipage}[t]{0.485\linewidth}
\textbf{Step 1: Window 1, positions $1:3$}\par\smallskip
\colorbox{exampleWake}{%
\begin{minipage}{\dimexpr\linewidth-2\fboxsep\relax}
\textbf{Wake}
\setlength{\topsep}{2pt}
\begin{tabbing}
Project $\h_{1:3}^{1}$ to $\Q_{1:3}^{1},\K_{1:3},\V_{1:3}$.\\
$\Ob_{1:3}^{1}\gets
  \softmax(\Q_{1:3}^{1}\K_{1:3}^\top/\sqrt d+\M)\V_{1:3}$\\
$\mathcal L\gets\mathcal L+
  \mathrm{CE}(\operatorname{Unembed}(\Ob_{1:3}^{1}),x_{2:4})$
\end{tabbing}
\end{minipage}%
}\par\smallskip
\colorbox{exampleSleep}{%
\begin{minipage}{\dimexpr\linewidth-2\fboxsep\relax}
\textbf{Sleep}
\setlength{\topsep}{2pt}
\begin{tabbing}
\hspace{0.85em}\=\kill
\textbf{for} $l=2,\ldots,N$ \textbf{do}\\
\> $\h_t^{l}\gets\ovec_t^{l-1}$ for $t=1,2,3$\\
\> Project $\h_{1:3}^{l}$ to $\Q_{1:3}^{l},
     \K_{1:3}^{\mathrm{new}},\V_{1:3}^{\mathrm{new}}$.\\
\> $\Ob_{1:3}^{l}\gets
     \softmax(\Q_{1:3}^{l}(\K_{1:3}^{\mathrm{new}})^\top/\sqrt d)
     \V_{1:3}^{\mathrm{new}}$\\
\> $\alphab_K\gets\sigma(
     [\K_{1:3}^{\mathrm{new}},\K_{1:3}]
     \thetab_{\alphab,K}^\top+b_{\alphab,K})$\\
\> $\alphab_V\gets\sigma(
     [\V_{1:3}^{\mathrm{new}},\V_{1:3}]
     \thetab_{\alphab,V}^\top+b_{\alphab,V})$\\
\> $\K_{1:3}\gets
     (1-\alphab_K)\odot\K_{1:3}^{\mathrm{new}}
     +\alphab_K\odot\K_{1:3}$\\
\> $\V_{1:3}\gets
     (1-\alphab_V)\odot\V_{1:3}^{\mathrm{new}}
     +\alphab_V\odot\V_{1:3}$\\
\textbf{end for}
\end{tabbing}
\end{minipage}%
}
\end{minipage}
\hfill
\begin{minipage}[t]{0.485\linewidth}
\textbf{Step 2: Window 2, positions $4:6$}\par\smallskip
\colorbox{exampleWake}{%
\begin{minipage}{\dimexpr\linewidth-2\fboxsep\relax}
\textbf{Wake}
\setlength{\topsep}{2pt}
\begin{tabbing}
Project $\h_{4:6}^{1}$ to $\Q_{4:6}^{1},\K_{4:6},\V_{4:6}$.\\
$\Ob_{4:6}^{1}\gets
  \softmax(\Q_{4:6}^{1}\K_{1:6}^\top/\sqrt d+\M)\V_{1:6}$\\
$\mathcal L\gets\mathcal L+
  \mathrm{CE}(\operatorname{Unembed}(\Ob_{4:6}^{1}),x_{5:7})$
\end{tabbing}
\end{minipage}%
}\par\smallskip
\colorbox{exampleSleep}{%
\begin{minipage}{\dimexpr\linewidth-2\fboxsep\relax}
\textbf{Sleep}
\setlength{\topsep}{2pt}
\begin{tabbing}
\hspace{0.85em}\=\kill
\textbf{for} $l=2,\ldots,N$ \textbf{do}\\
\> $\h_t^{l}\gets\ovec_t^{l-1}$ for $t=4,5,6$\\
\> Project $\h_{4:6}^{l}$ to $\Q_{4:6}^{l},
     \K_{4:6}^{\mathrm{new}},\V_{4:6}^{\mathrm{new}}$.\\
\> $\Ob_{4:6}^{l}\gets
     \softmax\!\left(
       \Q_{4:6}^{l}
       \begin{bmatrix}
         \K_{1:3}\\
         \K_{4:6}^{\mathrm{new}}
       \end{bmatrix}^{\!\top}\!/\sqrt d
     \right)
     \begin{bmatrix}
       \V_{1:3}\\
       \V_{4:6}^{\mathrm{new}}
     \end{bmatrix}$\\
\> $\alphab_K\gets\sigma(
     [\K_{4:6}^{\mathrm{new}},\K_{4:6}]
     \thetab_{\alphab,K}^\top+b_{\alphab,K})$\\
\> $\alphab_V\gets\sigma(
     [\V_{4:6}^{\mathrm{new}},\V_{4:6}]
     \thetab_{\alphab,V}^\top+b_{\alphab,V})$\\
\> $\K_{4:6}\gets
     (1-\alphab_K)\odot\K_{4:6}^{\mathrm{new}}
     +\alphab_K\odot\K_{4:6}$\\
\> $\V_{4:6}\gets
     (1-\alphab_V)\odot\V_{4:6}^{\mathrm{new}}
     +\alphab_V\odot\V_{4:6}$\\
\textbf{end for}
\end{tabbing}
\end{minipage}%
}
\end{minipage} 
\end{minipage}%
}
\endgroup
\caption{One-layer transformer forward pass with sleep.
}
\label{fig:sleep-example}
\end{figure*}

\section{Method}
\label{sec:method}
We adapt sleep~\citep{lee2026language} to attention-only Transformer models trained on natural language, with the aim of improving in-context sample efficiency.
\citet{lee2026language} let the model periodically revisits a window of context to refine its fast weight.
Since we use attention-only transformers, we instead refine their KV cache during sleep.
Unlike \citet{lee2026language}, we retain the full context without evicting tokens.
{We provide pseudocode for a toy example in Figure \ref{fig:sleep-example}.}

\paragraph{Wake--sleep cycle.}
During the wake phase, the model autoregressively decodes tokens with a single forward pass per token (\Cref{fig:decode}).
After every $W$ tokens, it sleeps: it performs $N-1$ additional depth-recurrent forward passes over the completed window to refine its KV cache.
Each token therefore receives $N$ forward passes in total, including the initial wake pass.
After sleep, decoding resumes at its normal speed until the next window is complete.

\paragraph{Training through sleep.}
\Cref{fig:arch} shows how this computation is organized for parallel training within each window.
The model is recurrent in depth and block-recurrent in time.
For each window of $W$ tokens, we apply the standard language modeling loss only to the first-loop output.
The remaining $N-1$ steps refine the cache without changing the current window's predictions.
Their effect on the loss comes from subsequent windows, whose predictions attend to the refined cache.
\Cref{fig:sleep-example} shows how the forward pass works during training for a 1-layer transformer model.
After forward pass, we backpropagate through the entire computation (including $N$ recursions) and update the model parameters.
Since refinement steps at $l\geq2$ do not need to obey causality within the current window, we use bidirectional attention during sleep.
In \Cref{fig:sleep-example}, we see that during sleep, the mask $\M$ is not applied to the attention.
During training, after the forward pass finishes, we backpropagate through the entire computation {(including recursions)} and update the model parameters.

\paragraph{Attention Residuals.}
To support signal propagation through depth recurrence, we use Attention Residuals (AttnRes)~\citep{team2026attention} with block size four.
Following \citet{su2026low}, we compute depth-wise attention weights using a subset of the hidden coordinates.
Rather than keeping the routing dimension constant across model sizes, we allocate $1/16$ of the hidden dimension to routing.
We select these architectural choices using the $N=1$ baseline, where they improve validation loss, and transfer them to the sleep models.

\paragraph{Gated cache refinement.}
Each refinement step produces candidate keys and values, which we combine with the previous cache through a GRU-style gated update.
Specifically, for the sleep window size $W$ and window index $i$, let $\K^{\mathrm{new}}_{(i-1)W+1:iW},\V^{\mathrm{new}}_{(i-1)W+1:iW}$
denote the candidates from the current step and
$\K_{(i-1)W+1:iW},\V_{(i-1)W+1:iW}$ the retained cache for the i-th window.
We first compute an elementwise retention gate (omitting window indices for readability):
\begin{align}
{\alphab}_K
=\sigma\!\left(
[\K^{\mathrm{new}},\K]\thetab_{\alphab,K}^\top+b_{\alphab,K}
\right),
\end{align}
where $\thetab_{\alphab,K}\in\mathbb{R}^{d_{\mathrm{KV}}\times 2d_{\mathrm{KV}}}$
and $b_{\alphab,K}\in\mathbb{R}^{d_{\mathrm{KV}}}$ are learned parameters,
$d_{\mathrm{KV}}$ is the total width of the KV heads,
$[\cdot,\cdot]$ denotes concatenation along the feature dimension,
and $\sigma$ is the sigmoid function.
We then update
\begin{equation}
\K\gets
(1-\alphab_K)\odot\K^{\mathrm{new}}
+\alphab_K\odot\K,
\end{equation}
where $\odot$ denotes elementwise multiplication.
We apply the same update to values using separate gate
parameters.
As shown in \Cref{fig:sleep-example}, this gating does \textit{not} affect the attention output in the current window, as it affects $\K,\V$ but not $\K^{\mathrm{new}},\V^{\mathrm{new}}$.
Unlike gating that modulates the forward pass activations~\citep{vendrell2026memory}, our gate controls what is retained in the cache independently of the forward-pass dynamics, which are controlled by AttnRes.
In \Cref{fig:kv_gating}, we show that adding a separate gating mechanism on top of AttnRes for retained KV cache is crucial.
The gating increases the number of parameters by less than 2\%.

\paragraph{Connection with looped transformers}
In \Cref{fig:sleep-example}, if we instead use final loop outputs $\Ob_{1:3}^N$ (with a causal mask enabled) and $\Ob_{4:6}^N$ to compute $\mathcal L$, this becomes block-recurrent looped transformers with KV cache sharing, similar to \citet{vendrell2026memory} but with our gating.
We call this \textit{block-recurrent loops}.
If we further make $W$ the same as the sequence length, we recover the standard looped transformers~\citep{dehghani2018universal} (gating does not affect the loss in this case).
We call this \textit{full-context loops}.
In \Cref{sec:ablation-baseline}, we compare our method against these baselines.

\section{Experimental setup}

\begin{table}[]
\centering
\caption{Model configurations for $N=1$, with tied embeddings.
$N>1$ adds approximately 2\% non-embedding parameters.
}
\label{tab:model-configs}
\scriptsize
\begin{tabular}{@{}rrrr@{}}
\toprule
Width ($d$) & Layers ($L$) & Non-embed. / total (M) & Batch (Mi tokens) \\
\midrule
512  & 6  & 22.93 / 48.10     & 0.5 \\
768  & 8  & 69.46 / 107.21    & 0.5 \\
1024 & 12 & 182.95 / 233.28   & 1 \\
1280 & 14 & 331.02 / 393.93   & 1 \\
1536 & 16 & 548.35 / 623.84   & 2 \\
2048 & 18 & 1096.20 / 1196.86 & 4 \\
\bottomrule
\end{tabular}
\end{table}

\paragraph{Pretraining and learning rates.}
We pretrain on DCLM~\citep{li2024datacomp} with a context length of 4,096 tokens.
We build upon the SmolLM-2~\citep{allal2025smollm2} architecture and tokenizer, varying width and depth following the Marin~\citep{marin} scaling ladder (\Cref{tab:model-configs}).
We use Kimi-style Muon~\citep{liu2025muon}, which ties the learning rates of the Muon and AdamW parameter groups.
Following Marin, we linearly warm up the learning rate over the first 1\% of optimizer updates, then linearly decay it to 5\% of its peak value over the remaining updates.
Learning rates are tuned for the $N=1$ baseline and reused for models with sleep.
In some of our sweeps, the best-performing learning rate
differs across loop counts, so this transfer may leave our sleep models undertuned.

We use labels such as 1.2B/32B to denote a configuration with 1.2B-parameter model trained on 32B tokens. We also interchangeably use d2048/TP30, where d2048 denotes a hidden width of 2,048 and TP30 denotes a pretraining budget of 30 tokens per non-embedding parameter.
See \Cref{app:additional-setup} for more details.

\begin{figure*}[t]
    \centering
    \includegraphics[width=1\linewidth]{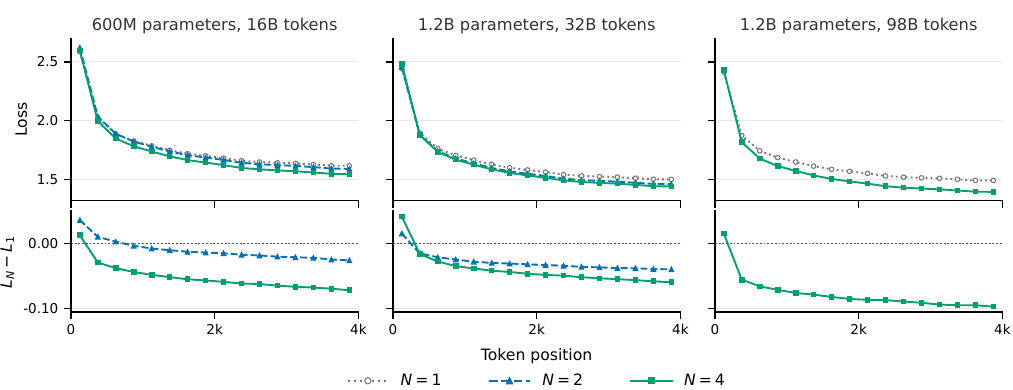}
    \caption{\textbf{Next-token prediction loss as context accumulates.}
The upper panels of each subplot show $L_N$, the loss for a model with $N$ loops.
Each point shows the mean next-token loss across evaluation sequences within a non-overlapping 256-token position bin.
Lower panels show $L_N-L_1$, the difference from the no-sleep baseline.
The evluation dataset combines arXiv2025 and Repo2025.
We show three configurations: d1536/TP30 (or 600M/16B), d2048/TP30 (or 1.2B/32B), and d2048/TP90 (or 1.2B/98B).
We observe similar trends on held-out ProLong dataset.
Full plot can be found in \Cref{fig:nll-full}.
} 
    \label{fig:nll_ood}
    \vspace{-5mm}
\end{figure*}

\section{Evaluation}

We start by presenting our main results on negative log-likelihood evaluations (\Cref{sec:nll}).
We then proceed to few-shot learning tasks, where the model should infer hidden rules from in-context demonstrations (\Cref{sec:few-shot}).
In \Cref{sec:additional-analyses}, we discuss additional experimental results including inference cost profile, ablation studies, and comparison with other looped baselines.

\subsection{Negative log-likelihood evaluation}
\label{sec:nll}

Our development was largely guided by next-token negative log-likelihood (NLL) evaluation.
Although multiple choices or generative evaluations such as MMLU or MATH are common,  they can be highly sensitive to minor details such as formatting and scoring rules~\citep{sclar2024quantifying}, and widely used ones are prone to contamination.
Moreover, benchmarks can be too hard for small models and thus fail to provide informative guidance at those scales~\citep{heineman2026signal}.
On the other hand, NLL evaluation provides a smooth signal, is less prone to minor variations such as formatting, and aligns well with downstream capabilities \cite{du2024understanding,gadre2025language}, especially when the evaluation corpus is carefully decontaminated and selected for quality.
In our setting, we leverage the fact that our training datasets, DCLM and ProLong, were released before 2025, to collect long-context data released in 2025 or later for testing purposes.
\textit{arXiv2025} contains arXiv papers released in 2025 and extracted into Markdown.\footnote{\url{https://huggingface.co/datasets/SlowGuess/Arxiv_2025_OCR}}
Additionally, we created \textit{Repo2025}, which contains 13 popular GitHub repositories whose earliest commits date to 2025.
These datasets let us test how well the model can adapt to scientific text and unfamiliar codebases; it is of course likely that both arXiv2025 and Repo2025 include AI-generated text, so our evaluation may not be fully uncontaminated.
Dataset construction is described in \Cref{sec:icl_tasks}.

\paragraph{NLL Results.}
\Cref{fig:nll_ood} plots loss against token position, averaged within 256-token windows.
Loss decreases as more context becomes available.
Our main finding is that beyond the first few hundred tokens, \textbf{models that are trained and tested with sleep achieve lower loss than the $N=1$ (no sleep) baseline;
the gap grows with context length, and longer sleep produces larger gains.}

The middle panel of \Cref{fig:main} shows this behavior at finer resolution, using a 10-token running mean.
The loss difference becomes more negative immediately after sleep boundaries, showing that sleep is the driving factor of the loss gap. The loss difference typically drops sharply at sleep boundaries and then rises within each window. We also sometimes observe similar trends in few-shot learning tasks (\Cref{fig:word-tasks-full}). We speculate that sleep pre-computes useful representations for upcoming tokens. Since near-future tokens are easier to anticipate, the model may learn to devote more computation to preparing for them, making the benefit largest immediately after sleep and smaller farther into the window.

How much context does a model need to reach the same prediction loss?
The right panel of \Cref{fig:main} shows loss on the same final 256 tokens of Repo2025 sequences as we vary the preceding context, using the d2048/TP30 models after 16k extension.
\textbf{$N=2$ and $N=4$ require $2.10\times$ and $2.39\times$ fewer prefix tokens, respectively, to match $N=1$'s measured loss at its longest context.}

\subsection{Few-shot learning tasks}
\label{sec:few-shot}
\paragraph{Setup.}
Although the NLL evaluation provides a robust predictor of in-context adaptation ability,  we additionally evaluate on several few-shot learning tasks to complement our NLL results.
\textit{CoQA}~\citep{reddy2019coqa} is a standard benchmark~\citep{radford2019language,brown2020language,chowdhery2023palm} where the model answers questions about a passage through a conversation.
Unlike other evaluations, this is a generative evaluation.
Here, we use greedy decoding and report F1.
For 1-, 2-, and 4-shot results, we average the scores over two fixed demonstration sets shared across models.
\textit{Algorithmic word tasks} test whether sleep improves models' ability to infer the correct algorithmic rule from in-context demonstrations and apply it correctly.
We consider five tasks: word reversal, cyclic rotation, selective recall, symbol removal, and latest assignment.
For character manipulation tasks, we separate characters with spaces to reduce dependence on word tokenization.
We report the probability of the correct answer, computed from teacher-forced token probabilities, averaged over all subtasks.
We also evaluate on three tasks from the widely used \citep{abdin2024phi,team2025kimi,zeng2026glm} ICL subset of \textit{HELMET} dataset~\citep{yen2025helmet}: TREC-coarse, TREC-fine~\citep{li2002learning}, and NLU~\citep{liu2021benchmarking}.
These are text classification tasks with 6, 50, and 68 labels, respectively, for TREC-coarse, TREC-fine, and NLU.
For each sequence, we randomly assign numeric labels to classes and keep the mapping fixed throughout, so the model can infer the label meanings from in-context demonstrations.
As in the algorithmic tasks, we report the probability of the correct answer averaged over all subtasks.
Finally, we evaluate in-context sentiment classification on \textit{SST-2}, replacing
\texttt{negative} and \texttt{positive} with \texttt{foo} and
\texttt{bar}, respectively, following \citet{wei2023larger}.
These labels do not reveal their sentiment meanings, so the model must
learn the mapping from demonstrations.
Learning new association solely from context can be surprisingly hard: \citet{wei2023larger} shows that even GPT-3 achieves around 60\% accuracy with 32 demonstrations.
For a detailed explanation of the tasks used, see \Cref{sec:icl_tasks}.

\begin{figure}
    \centering
    \includegraphics[width=1.\linewidth]{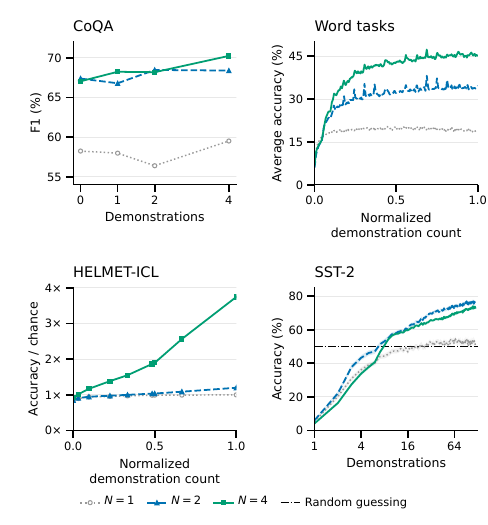}
    \caption{\textbf{Few-shot learning performance as demonstrations accumulate.} We evaluate on four few-shot learning tasks, CoQA, algorithmic word tasks, HELMET-ICL, and SST-2. The d2048/TP30 (or 1.2B/32B) model is used, where we see longer sleep duration leads to significant improvements as the number of demonstration increases. Other configurations follow similar trends (\Cref{app:additional-few-shot}). }
    \label{fig:few-shot-main}
\end{figure}

\paragraph{Few-shot learning results.}
\Cref{fig:few-shot-main} presents the results for d2048/TP30 (or 1.2B/32B). These results show that \textbf{models with sleep outperform the $N=1$ baseline on few-shot learning across all four datasets.}
In \Cref{app:additional-few-shot}, we find that the similar trend holds for d1536/TP30 (or 600M/16B) and d2048/TP90 (or 1.2B/98B) configurations.
For CoQA, the advantage is also present without prepended demonstrations because even zero-shot CoQA prompts can exceed the first sleep boundary at 256 tokens.
In algorithmic tasks, the $N=1$ baseline often plateaus, while models with sleep continue to improve as demonstrations accumulate.
In SST-2, the $N=1$ baseline's gains diminish as context
accumulates, while models with sleep continue to benefit from
additional examples.
All models struggle to even achieve 50\% accuracy initially, showing that even identifying a valid answer set in-context can be challenging for small models on this task.
We sometimes see that $N=2$ is tied with $N=4$ or even outperforms it, e.g., in CoQA and SST-2, at certain demonstration counts.
This is consistent with the fact that for d2048/TP30, the gap between $N=2$ and $N=4$ is smaller than that between $N=1$ and $N=2$ on the NLL evaluation, and that an individual few-shot learning task does not necessarily reflect a model’s general ability to adapt in context.

\begin{figure}[t]
  \centering
  \includegraphics[width=1\linewidth]{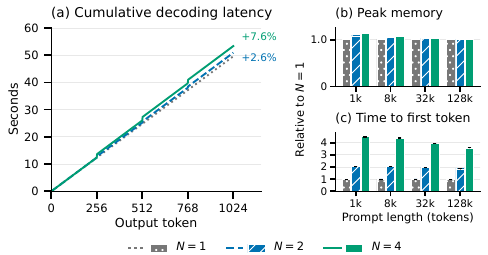}
  \caption{Inference cost profile comparison between $N=1,2,4$:
  (a) cumulative decoding latency, excluding time to first token.
  (b) peak memory allocation,
  (c) time to first token relative to $N=1$.
  All panels are measured using a 1.2B model with $W=256$, batch size 8,
  and 1,024 generated tokens on an RTX 6000 Ada GPU.  We see that a model with longer sleep incurs a proportionally higher time to first token, but otherwise incurs no additionally overhead during wake-time decoding, and only minimally increases overall memory footprint and cumulative latency at decoding stage.}
  \label{fig:inference}
\end{figure}

\subsection{Additional analyses}
\label{sec:additional-analyses}

\paragraph{Inference cost profile.}
\label{sec:inference}
\Cref{fig:inference} measures inference cost for $N=1,2,4$ using
d2048 (or 1.2B) models with $W=256$.
We use BF16 weights, batch size 8, and CUDA graphs for decoding
on an RTX 6000 Ada GPU.

\Cref{fig:inference}(a) shows that sleep adds no decoding
latency between sleep boundaries.
The additional latency is concentrated in periodic spikes,
when the model refines the completed window's KV cache.
Sleep processes the window's tokens in parallel, and its cost
is amortized over subsequent decoding steps.
As a result, with a 128k token prefix,
end-to-end cumulative decoding time (excluding time to first token) increases by only 2.6\% for $N=2$ and 7.6\% for $N=4$.
\Cref{fig:inference}(b) shows that memory overhead is also small: at the longest prompt length, peak memory increases by about 1\% for $N=4$ (mainly due to AttnRes cache and KV gating).
The main latency trade-off is prefilling: time to first token scales approximately with $N$,
as shown in \Cref{fig:inference}(c).
This additional computation lets the model process its
context more deeply before generating a response, improving
the sample efficiency of in-context adaptation.

\paragraph{Training overhead.}
As with other looped transformer baselines, our model incurs approximately $N\times$ training FLOPs over the no-sleep baseline (KV gating FLOPs are negligible).
The sleep window size $W$ does not change the FLOPs, but it affects parallelism.
Using too small of a $W$ can lead to poor GPU utilization, so unless otherwise mentioned, we use $W=256$ to ensure that training is FLOPs-bound.

\begin{figure}
    \centering
    \includegraphics[width=1\linewidth]{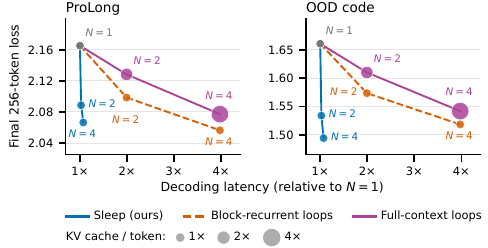}
    \caption{Loss on the final 256 tokens of 4k-token sequences
for 230M-parameter models on ProLong and Out-of-distribution code datasets. $N$ is the number of loops. Our models achieve comparable loss to looped transformer baselines (see \Cref{sec:method} for descriptions) with better latency and memory footprint. 
Decoding time is the cumulative latency for generating 1k tokens
after a 128k-token prefix (excluding prefill).
Circle area is proportional to KV-cache size per token.
}
\vspace{-5mm}

    \label{fig:baseline_comp_main}
\end{figure}

\paragraph{Comparison with other looped models.}
\label{sec:ablation-baseline}
We compare our method against two looped baselines we discussed in \Cref{sec:method}: block-recurrent loops and full-context loops.
For that, we train d1024/TP30 models with $N=1,2$, and $4$.

\Cref{fig:baseline_comp_main} shows the final 256-token losses as a function of the cumulative decoding time relative to $N=1$ on ProLong and Repo2025.
Unlike both looped baselines, which require $N$ steps to generate each token, our models do not incur wake time latency overhead and instead hide that additional FLOPs to sleep phases. 
\textbf{Since sleep is FLOPs intensive, it achieves superior cumulative decoding latency compared to these looped baselines which incur $N$ times longer latency.}
Moreover, sleep could be flexibly scheduled during otherwise idle periods, such as while awaiting tool results or user input, without harming interactivity.

\Cref{fig:readout-icl} compare the in-context adaptation curves of these methods.
We can see that full-context loops start from a lower position than our method, but eventually fall behind it, despite having $N\times$ larger KV cache and being much slower in decoding.
We believe this is because full-context loops do not use block-recurrence and thus their computational depth do not grow in temporal dimension.
Block-recurrent loops show steeper learning curves than full-context loops and outperform sleep at most token positions at the cost of increased decoding latency, but sleep eventually outperforms it at later positions.
We hypothesize that this is because sleep allows recurrent computation to focus on preparing the KV cache for predicting distant future tokens, whereas block-recurrent loops must use recurrence for immediate next-token prediction as well as cache preparation, potentially creating competition between these two roles~\citep{monea2026state}.
\textbf{This result is surprising because these baselines incur approximately $3.7\times$ more decoding latency, and full-context loops costs $4\times$ more KV cache memory on top of that (see \Cref{fig:baseline_comp_main}); nevertheless, when there is enough context for the model to adapt, sleep appears to outperform them eventually.}

\begin{figure}[]
    \centering
    \includegraphics[width=\linewidth]{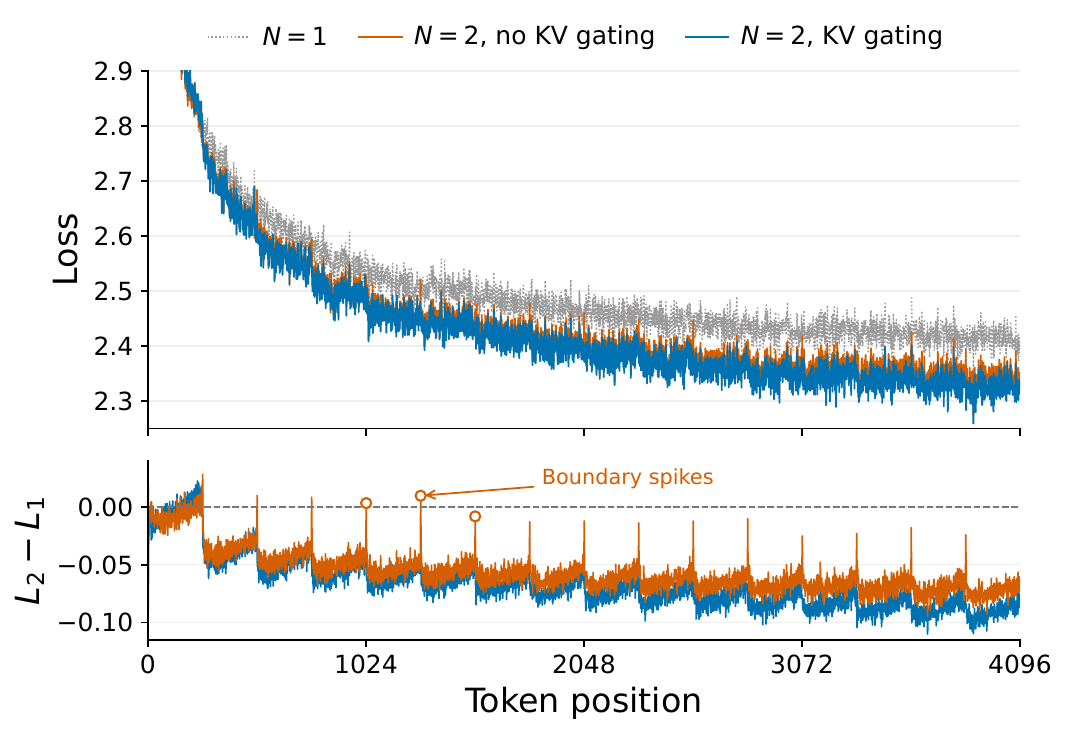}
    \caption{\textbf{Loss comparison, with and without KV gating.} Circles highlight sharp boundary spikes without gating; learned KV write gates substantially reduce them. d768/TP30 models are used.}
    \label{fig:kv_gating}
\end{figure}

\begin{figure*}[]
    \centering
    \includegraphics[width=\linewidth]{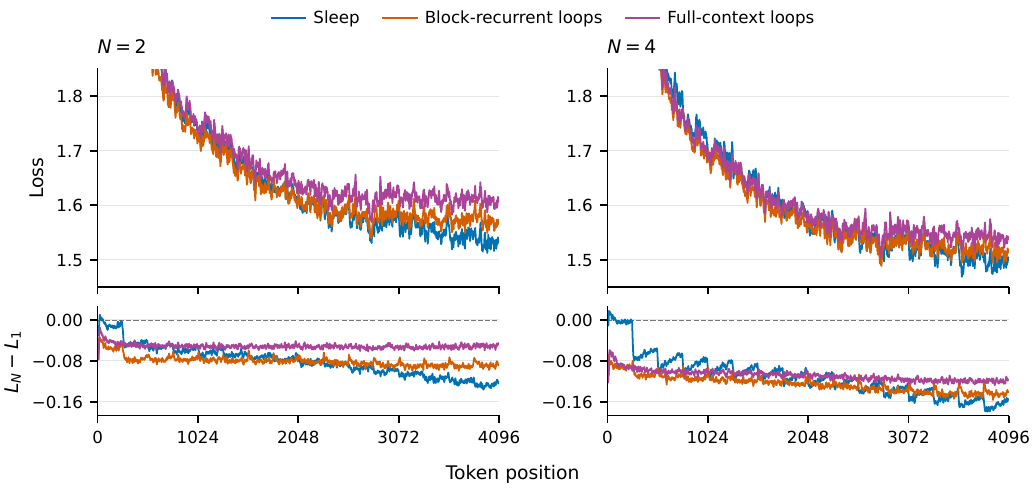}
   \caption{\textbf{Learning curve comparison for sleep, block-recurrent
loops, and full-context loops.}
Loss per token is measured on Repo2025 using d1024/TP30 models
with $N=2$ (left) and $N=4$ (right).
Upper panels show loss, and lower panels show the loss difference from the shared $N=1$ baseline.
}
    \label{fig:readout-icl}
\end{figure*}

\paragraph{KV gating.}
KV gating separates what the model computes during sleep from what it
retains for subsequent windows.
Later loops can develop features useful for refinement without forcing
those features to replace the existing cache.
Without gating, the final-loop KV overwrites earlier representations,
potentially displacing low-level features that subsequent windows may need.

Although AttnRes could limit cache changes by letting the query vector
attend to previous-loop representations rather than current-loop
updates, this also constrains the features being generated.
Consistent with this interpretation, the lower panel of \Cref{fig:kv_gating} shows that boundary spikes persist without KV gating even when AttnRes is used.
\textbf{Gating suppresses these spikes and improves loss at later positions.}
A useful analogy is the observation that people often cannot recall REM sleep~\citep{crick1983function}, implying that it is useful to have the separation of generating neural activities from what is retained in memory.

\begin{figure}[]
    \centering
    \includegraphics[width=0.88\linewidth]{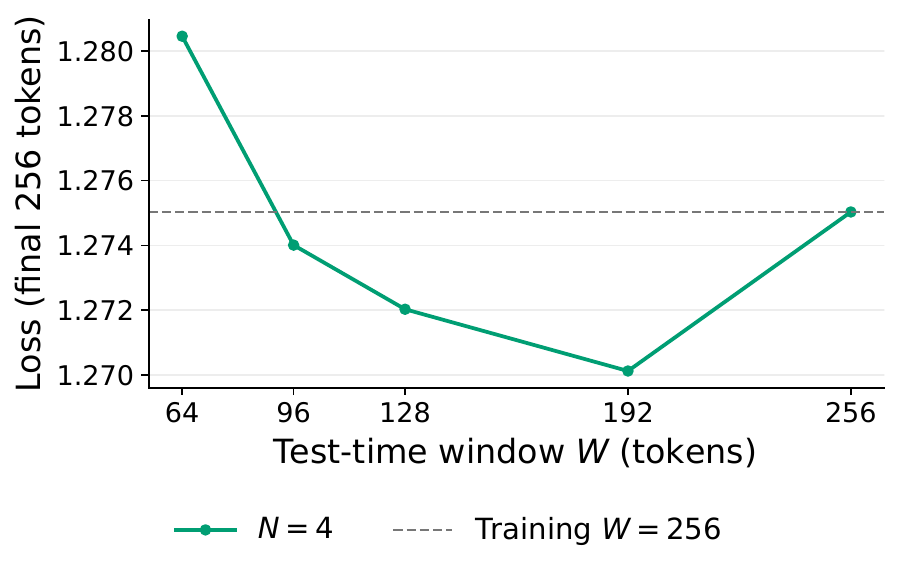}
    \caption{\textbf{Changing the sleep window at test time.}
    Repo2025 loss on the final 256 tokens of 4,096-token sequences
    for the d1536/TP30 model with $N=4$, trained with $W=256$.
    $W=192$ achieves the lowest loss among the tested windows.}
    \label{fig:test-time-window}
\end{figure}

\paragraph{Changing $W$ during inference.}
\Cref{fig:test-time-window} shows that {the model tolerates moderate
changes to $W$ at test time, with some smaller windows even improving
loss without retraining.}
This flexibility may partly arise from document packing during training:
if attention largely ignores tokens before the preceding \texttt{<eos>},
the model effectively learns to refine within-document segments
shorter than $W$.
Randomizing $W$ during training could further support flexible sleep
schedules, allowing the model to sleep while awaiting tool results
or user input and thereby reduce interruptions during interaction.
We leave such adaptive training and scheduling strategies to future work.

\paragraph{Other ablations.}
In \Cref{sec:app:additional-exp}, we present additional experiments on standard tasks from \texttt{lm-eval-harness}, recurrence layout, effect of bidirectional attention during sleep, and window size ablation.

\section{Conclusion}
We study periodic context refinement via depth recurrence, or sleep, as a way to improve
in-context sample efficiency of transformer language models without increasing model size.
Across NLL and few-shot evaluations, we show that models with sleep make more effective use of their context
and that longer sleep duration $N$ improves results further.

Our experiments are \textit{not} iso-FLOPs.
We instead match training tokens, and sleep requires approximately $N$ times the training computation of the no-sleep baseline per training token.
This trade-off is attractive when inference demand is large enough to amortize the additional training cost.
Also, our method makes prefill approximately $N$ times slower.
Therefore, we expect the benefit to be the greatest in a decoding-heavy scenario.

When a model has enough context and time to adapt, the slope of its learning curve  may matter more than what it already knows. Our results strongly suggest that the performance gap between models with and without sleep will continue to widen as the context accumulates. Future work should scale this work to larger, more capable models to investigate if this trend persists at longer context and after post-training, and by investigating other opportunities for language models to improve on in-context adaptation.

\section*{Acknowledgement}
We thank Zhengyang Geng for valuable contribution to the initial phase of this project during his time at CMU, Sean McLeish for comments on the draft, and Yelong Shen for discussions.

\bibliography{reference}
\bibliographystyle{icml2026}

\newpage
\appendix
\onecolumn

\section{Additional setup, analyses, and discussions}
\label{sec:app:additional-exp}

\subsection{Additional experimental setup}
\label{app:additional-setup}

We build upon the SmolLM-2~\citep{allal2025smollm2} architecture and tokenizer, varying width and depth following the Marin~\citep{marin} scaling ladder.
The models are dense, global-attention only Transformers with grouped-query attention~\citep{ainslie2023gqa} and SwiGLU activations~\citep{shazeer2020glu}.
We tie the embedding and unembedding weights, use NoPE~\citep{kazemnejad2023impact} in every fourth layer, and include Inkling-style residual short convolutions~\citep{thinkingmachines2026inkling}.

To support signal propagation through depth recurrence, we use Attention Residuals~\citep{team2026attention} with block size four.
Following \citet{su2026low}, we compute depth-wise attention weights using a subset of the hidden coordinates.
Rather than keeping the routing dimension constant across model sizes, we allocate $1/16$ of the hidden dimension to routing.
We select these architectural choices using the $N=1$ baseline, where they improve validation loss, and transfer them to the sleep models.
\Cref{tab:short} shows that these combined result in improved performance on the standard \texttt{lm-eval-harness} tasks compared to an external Transformer++ baseline.
Unless otherwise stated, models with sleep recur the full
Transformer stack except for embedding and unembedding layers every $W=256$ tokens.

Following Marin's fitting approach, we obtain the heuristic
$$\eta = 0.97082\ T^{-0.308625}\,d^{-0.687709}\,B^{0.5},$$
where $\eta$ is the learning rate, $T$ is the total number of training tokens, $d$ is the hidden dimension, and $B$ is the global batch size in tokens.
The fit uses sweeps for d512, d768, and d1024 at TP30, TP60, TP150, and TP300, together with sweeps for d1280/TP30 and d1536/TP30.
We extrapolate this heuristic to d2048/TP90.

In our initial d2048/TP90 experiments, we observed large RMS magnitudes in activations.
To control their scale, we add normalization after both the attention and MLP outputs while retaining the existing pre-normalization, yielding sandwich normalization.
We also set the peak learning rate to $0.7\times$ the heuristic prediction.
Both $N=1$ and $N=4$ in the reported TP90 comparison use these modifications.
We were unfortunately unable to run $N=2$ for d2048/TP90 because our compute access was terminated.

\paragraph{Long-context extension.}
For the d2048/TP30 checkpoints, we extend the context length to 16,384 tokens through an additional 3.3B tokens of training using YaRN~\citep{peng2024yarn}.
We  linearly decay the learning rate from $0.05\times$ to $0.01\times$ the base learning rate during this stage.
This stage uses a mixture of 70\% DCLM and 30\% ProLong~\citep{gao2025train}.
From ProLong, we use the \texttt{thestackv1}, \texttt{book}, and \texttt{textbooks} subsets to obtain contiguous sequences.

\begin{table*}[t]
\centering
\caption{Zero-shot downstream evaluation on the standard \texttt{lm-eval-harness} tasks.
Accuracy is reported in percent; * denotes length-normalized accuracy.
Average is the unweighted mean of the six accuracy metrics,
excluding LAMBADA perplexity.
}
\label{tab:short}
\scriptsize
\setlength{\tabcolsep}{6pt}
\begin{tabular}{@{}lrrrrrr@{}}
\toprule
& \multicolumn{3}{c}{Sleep (Ours): $\approx$1.2B, 32.89B tokens}
& \multicolumn{2}{c}{Sleep (Ours): $\approx$1.2B, 98.66B tokens}
& Transformer++ \\
\cmidrule(lr){2-4}\cmidrule(lr){5-6}
Metric
& $N=1$ & $N=2$ & $N=4$
& $N=1$ & $N=4$
& 1.3B, 100B tokens \\
\midrule
LAMBADA PPL $\downarrow$
& \textbf{6.21} & 6.37 & 6.43 & \textbf{5.25} & 5.56 & 18.32 \\
LAMBADA
& \textbf{60.20} & 59.17 & 59.29 & \textbf{63.90} & 62.93 & 42.60 \\
HellaSwag*
& 58.23 & 58.63 & \textbf{58.71} & \textbf{64.25} & 62.65 & 50.23 \\
PIQA
& 73.83 & \textbf{74.05} & 73.56 & \textbf{74.92} & 74.76 & 70.02 \\
ARC-E
& \textbf{68.01} & 66.20 & 67.47 & \textbf{70.88} & 70.66 & 68.83 \\
ARC-C*
& \textbf{33.28} & 32.59 & 32.34 & \textbf{37.54} & 35.41 & 35.10 \\
WinoGrande
& 60.14 & \textbf{61.25} & 59.91 & \textbf{62.35} & 61.09 & 53.51 \\
\midrule
Average $\uparrow$
& \textbf{58.95} & 58.65 & 58.55 & \textbf{62.31} & 61.25 & 53.38 \\
\bottomrule
\end{tabular}
\end{table*}

\begin{table}[]
    \centering
    \footnotesize
    \caption{ProLong validation loss for d1024/TP300 models trained with a
    4k-token context window. Layout denotes prelude--recurrent layers $\times$
    loops--coda. $0$--$12$--$0$ is the no-sleep baseline.
    Bold denotes the layout with lower loss
    for each logical depth.}
    \begin{tabular}{lccc}
        \toprule
        Layout & Avg. loss & Last-1k loss & Logical depth \\
        \midrule
        $0$--$12$--$0$ & 2.084 & 1.971 & 12 \\
        \midrule
        $4$--$4\times3$--$4$ & 2.057 & 1.940 & 20 \\
        $\mathbf{2}$--$\mathbf{8\times2}$--$\mathbf{2}$
            & \textbf{2.053} & \textbf{1.934} & 20 \\
        \midrule
        $2$--$8\times4$--$2$ & 2.043 & 1.925 & 36 \\
        $\mathbf{0}$--$\mathbf{12\times3}$--$\mathbf{0}$
            & \textbf{2.035} & \textbf{1.910} & 36 \\
        \bottomrule
    \end{tabular}
    \label{tab:layout-ablation}
\end{table}

\subsection{Short-context tasks}

\Cref{tab:short} shows short-context evaluation results.
All evaluated tasks are shorter than $W=256$ tokens, so models do not use any sleep.
As expected, results are similar to the baseline (if anything, models with sleep obtain slightly lower accuracies).
\textbf{This shows that indeed advantage of sleep comes from better in-context adaptation, not from the model already memorizing more knowledge or skills than the baseline.}
Transformer++ results are from \citet{yang2025gated}; this model is trained on 100B FineWeb-Edu tokens.
Our models achieve better performance likely because of several differences, such as Muon, AttnRes, and heavy use of short convolution layers.
These improvements allow us to conduct meaningful experiments even at small scales.

\subsection{Recurrence layout}
We first compare which layers to recur while holding logical depth fixed.
\Cref{tab:layout-ablation} shows that recurring more layers with fewer
loops achieves lower ProLong loss.
At logical depth 20, recurring eight layers twice outperforms
recurring four layers three times.
At logical depth 36, recurring the entire network three times
outperforms recurring only the middle eight layers four times.
These results motivate our choice to recur the entire network.

This is contrary to findings in depth-recurrence literature, where looping the middle part achieves better performance
in iso-FLOPs comparison~\citep{saunshi2025reasoning,wang2026smelt} and is thus a standard practice~\citep{geiping2026scaling,mcleish2025teaching,schwethelm2026much}.
We hypothesize that this difference arises because, unlike looped models
that retain separate KV caches for different loops, sleep repeatedly
refines the same cache.
Information from successive sleep recurrence must therefore share a fixed
cache capacity, potentially creating a trade-off between preserving
earlier information and incorporating later refinements.
At the same logical depth, a larger recurrent block provides more space for sleep to retain and refine information across loops.

\begin{table}[]
    \centering
    \caption{Causal-masking ablation for d1536/TP30 models with
    $N=2$, $W=256$, and 4k training context.
    The causal mask is varied only during sleep.}
    \label{tab:causal-ablation}
    \small
    \setlength{\tabcolsep}{4pt}
    \begin{tabular}{@{}lccc@{}}
        \toprule
        & \multicolumn{2}{c}{ProLong}
        & DCLM \\
        \cmidrule(lr){2-3}
        Causal mask during sleep
        & Avg. & Last-1k & Avg. \\
        \midrule
        With
        & 1.978 & 1.861 & 2.687 \\
        Without
        & \textbf{1.977} & \textbf{1.858} & \textbf{2.680} \\
        \bottomrule
    \end{tabular}
\end{table}

\subsection{Bidirectional attention during sleep}
During sleep, self-attention does not need to be causal because refinement steps do not contribute to predictions for tokens in the current window.
We compare causal and bidirectional attention during sleep.
\Cref{tab:causal-ablation} shows that bidirectional refinement
achieves slightly lower loss on all three validation metrics.

\begin{table}[t]
    \centering
    \caption{Window-size ablation for d768/TP30 models with $N=2$.}
    \label{tab:window-ablation}
    \small
    \begin{tabular}{@{}rrrr@{}}
        \toprule
        $W$ & DCLM train & ProLong avg. & Last-1k \\
        \midrule
        256  & \textbf{3.1592} & \textbf{2.4330} & \textbf{2.3062} \\
        512  & 3.1714 & 2.4626 & 2.3404 \\
        1024 & 3.1772 & 2.4698 & 2.3489 \\
        \bottomrule
    \end{tabular}
\end{table}

\subsection{Window size ablation}
\Cref{tab:window-ablation} shows the benefit of using smaller $W$.
Decreasing $W$ makes sleep more frequent, increasing the computational depth to process a sequence.
However, reducing $W$ also reduces the number of tokens processed in parallel and thus can negatively impact the model FLOPs utilization (MFU).
Therefore, we mostly use $W=256$ in our experiments so the forward pass can remain close to being FLOPs-bound.

\subsection{Few-shot learning tasks}
\label{app:additional-few-shot}
\Cref{fig:nll-full,fig:coqa,fig:word-tasks-full,fig:helmet-full} show results on NLL, CoQA, algorithmic word tasks, and HELMET-ICL, respectively for three different configurations: d1536/TP30, d2048/TP30, and d2048/TP90.
We see that
\textbf{models with sleep improve faster as demonstrations accumulate.}

\subsection{Discussion on test-time training}
\begin{figure}
    \centering
    \includegraphics[width=0.5\linewidth]{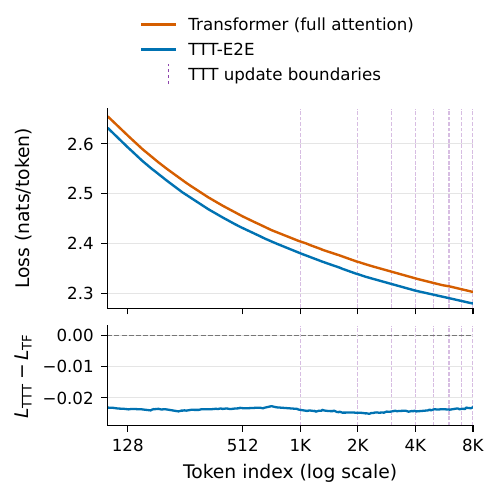}
    \caption{Plot extracted from Figure 6 of \citet{tandon2025end}. The top panel shows the next-token loss per token position, comparing their Transformer baseline and the proposed TTT-E2E. The bottom panel shows loss differences. Vertical dashed lines mark test-time updates every 1k tokens.}
    \label{fig:ttt-fig6}
\end{figure}

\begin{figure}
    \centering
    \includegraphics[width=1\linewidth]{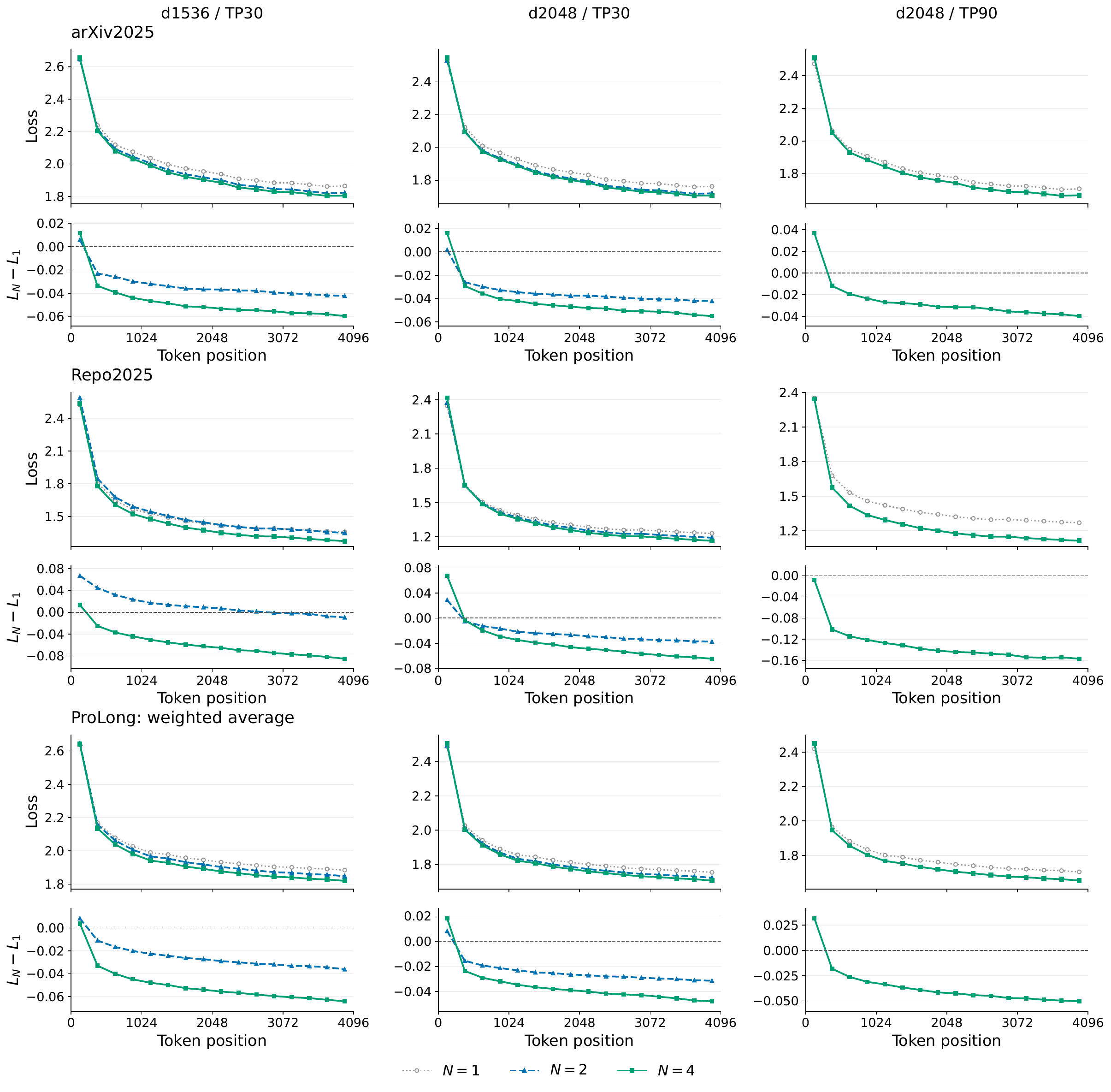}
    \caption{\textbf{Next-token prediction loss as context accumulates.}
Upper panels show $L_N$, the loss for a model with $N$ loops,
averaged within 256-token windows.
Lower panels show $L_N-L_1$, the difference from the no-sleep baseline.}
\label{fig:nll-full}
\end{figure}

\begin{figure*}
    \centering
    \includegraphics[width=0.85\linewidth]{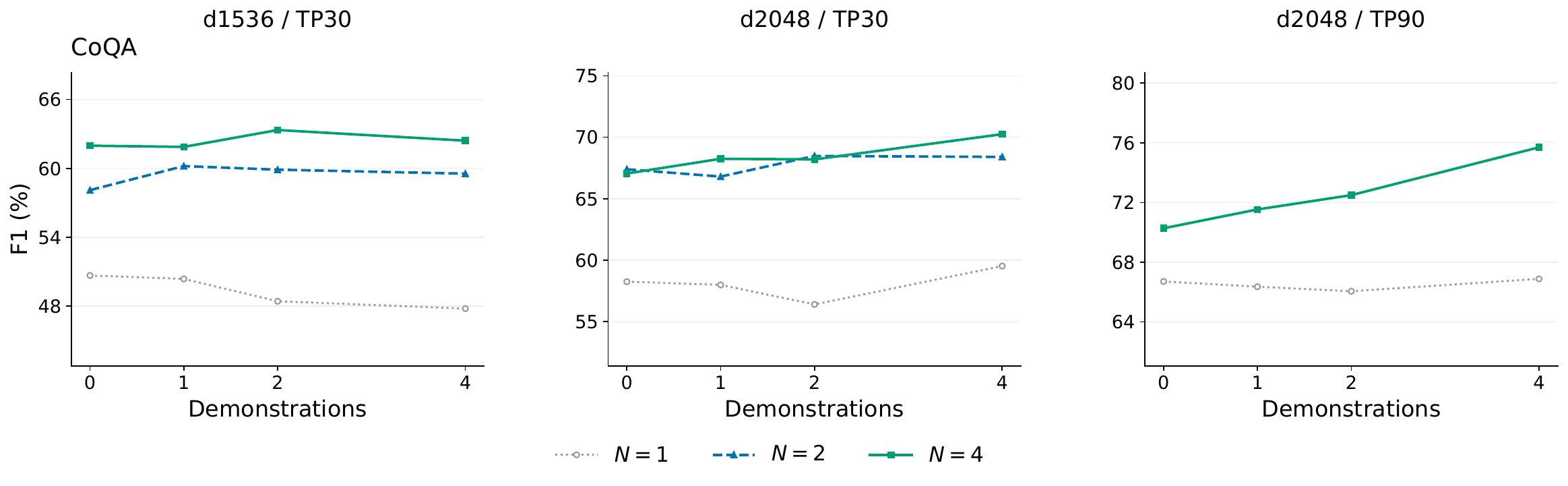}
    \caption{\textbf{Few-shot CoQA.}}
    \label{fig:coqa}
\end{figure*}

\begin{figure}
    \centering
    \includegraphics[width=0.85\linewidth]{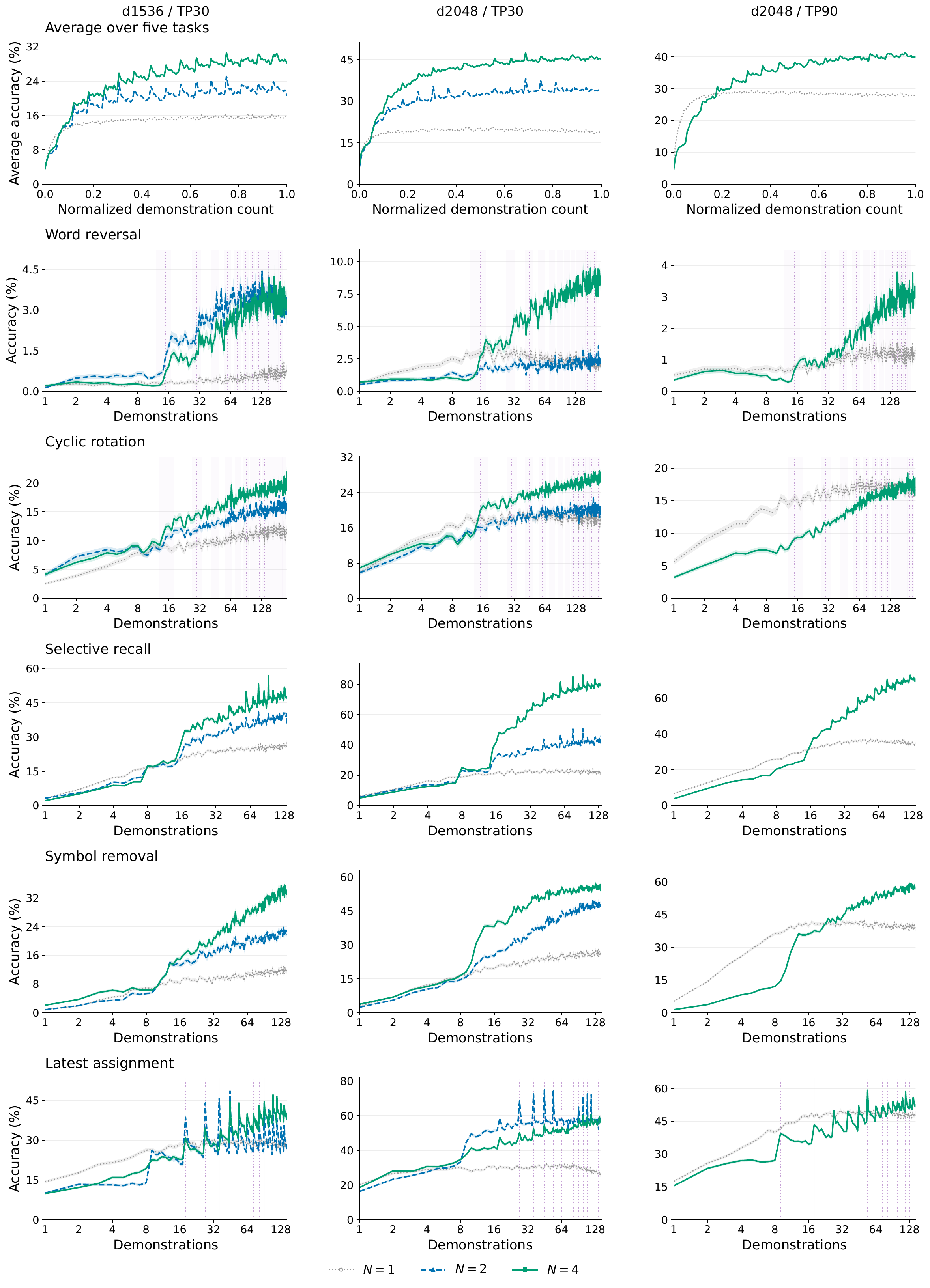}
    \caption{\textbf{In-context adaptation on five algorithmic word tasks.} The top row shows the average accuracy over all tasks, where x-axis is normalized between 0 and 1. 
    Vertical dotted lines show median sleep boundaries, and the shaded areas around them represent the 1st--99th percentile range (as boundaries can occur at different demonstration counts across examples). Sleep boundaries are omitted for the plots that average over multiple formats (selective recall and symbol removal).
    }
    \label{fig:word-tasks-full}
\end{figure}

\citet{tandon2025end} propose a test-time training method called TTT-E2E where the model performs gradient descent on next-token loss within context.
TTT-E2E originally aims to preserve the model's long-context ability after sliding-window token eviction, but since its sliding window size is 8k and test-time update happens every 1k tokens, we can also examine its loss improvement over the Transformer baseline before eviction to see if it can improve in-context adaptation.
In \cref{fig:ttt-fig6}, we show the plot digitized from Figure 6 (left) of \citet{tandon2025end}, where we display loss curves of their Transformer baseline and TTT-E2E. 
We restrict the x-axis to positions up to 8K, before token eviction.
TTT-E2E improves loss over the baseline, but the gap remains approximately constant over this range, unlike our \Cref{fig:main,fig:nll_ood}, where the gap grows as context accumulates.
We also see that  the advantage is already present before the first test-time update.
Why does this happen?
This comparison between transformer and TTT-E2E in \cref{fig:ttt-fig6} is not matched in terms of the size of per-sequence inference state, because unlike sleep, which updates the same KV cache that is used by a vanilla transformer, TTT-E2E introduces additional context-specific fast-weight states that are learned from context tokens in a sequence and thus cannot be shared across other independent sequences in a batch.
Having larger per-sequence inference state increases decoding latency through additional memory traffic because decoding is already memory-bound, while also improving in-context adaptation ability as noted by the authors.
For example, in their 3B parameter model, we estimate their in-context memory overhead (size of KV cache and fast weights combined) to be more than $1.5\times$ that of the  Transformer baseline, which has only the KV cache.
We obtain this estimate as follows.
Their 3B model has 32 layers, and they update the final 8 layers' SwiGLU parameters in-context. Their SwiGLU hidden and intermediate widths are 2,560 and 5,632, respectively, and the updated state is stored in 32-bit precision.
So in total, their SwiGLU state occupy $8 \cdot 5632 \cdot 2560 \cdot 3 \cdot 4$ bytes, or approximately 1.29GB.
On the other hand, their KV cache is stored in 16-bit precision, and since they do not use GQA, the size is $32 \cdot 2560 \cdot 2 \cdot 2 \cdot 8192=$2.5GB for 8k context.
So TTT-E2E incurs roughly 51.6\% larger per-sequence inference state size at 8k. 
Moreover, TTT-E2E only performs a single in-context gradient step, whereas we show that increasing sleep duration leads to better performance.

Why does sleep benefit from accumulated context in loss whereas TTT-E2E does not?
One reason could be that sleep can be seen as a form of black-box meta learning method~\citep{hochreiter2001learning,kirsch2022general,duan2016rl,wang2016learning}, where the model can execute flexible learned updates to \textit{train} the KV cache without relying on a prescribed inner-loop loss function or update rule.
On the other hand, TTT-E2E relies on prescribed inner loss (next-token loss) and update rule (gradient descent), which makes it more similar to MAML~\citep{finn2017model} than the black-box meta-learning approaches and may limit the flexibility of its updates when learning from few examples.
Aligned with this intuition, \citet{eyuboglu2026cartridges} find that using next-token loss as an inner loss for TTT can improve memorization but does not benefit other tasks such as mathematical reasoning as much.

\begin{figure*}
    \centering
    \includegraphics[width=0.85\linewidth]{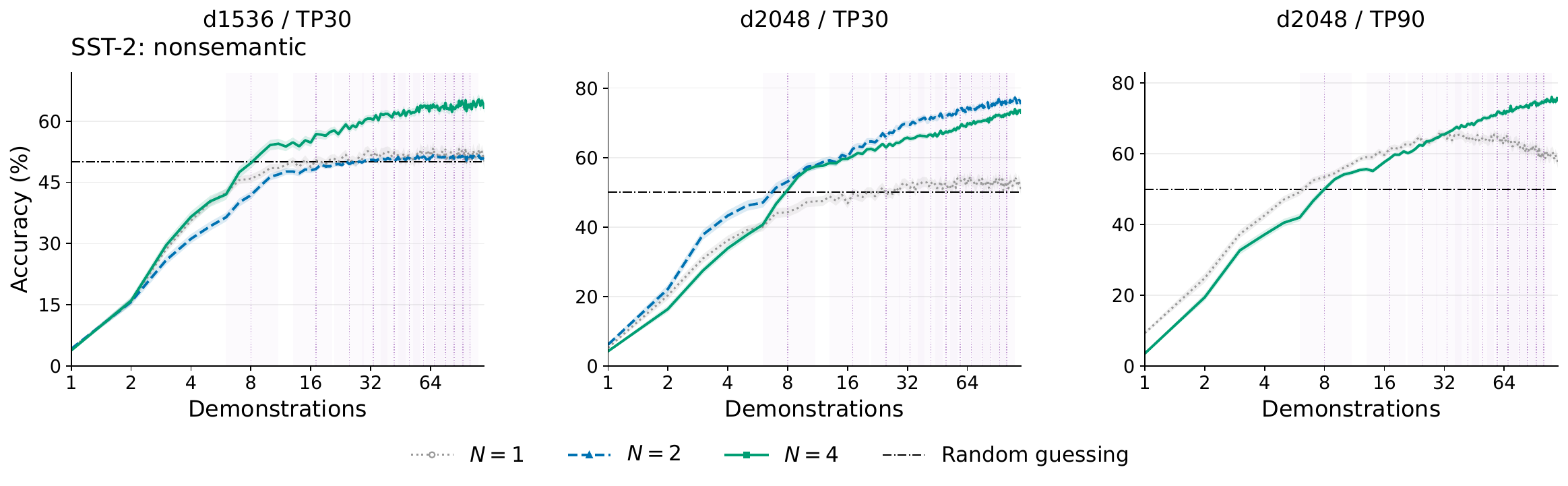}
    \caption{\textbf{SST-2, where instead of using positive/ labels, the labels are replaced by $\{\text{foo, bar}\}$}, so the model must learn the correct correspondence between these placeholders and the correct semantic labels via in-context demonstrations.  Vertical dotted lines show median sleep boundaries, and the shaded areas around them represent the 1st--99th percentile range.}
    \label{fig:sst}
\end{figure*}

\begin{figure*}
    \centering
    \includegraphics[width=0.85\linewidth]{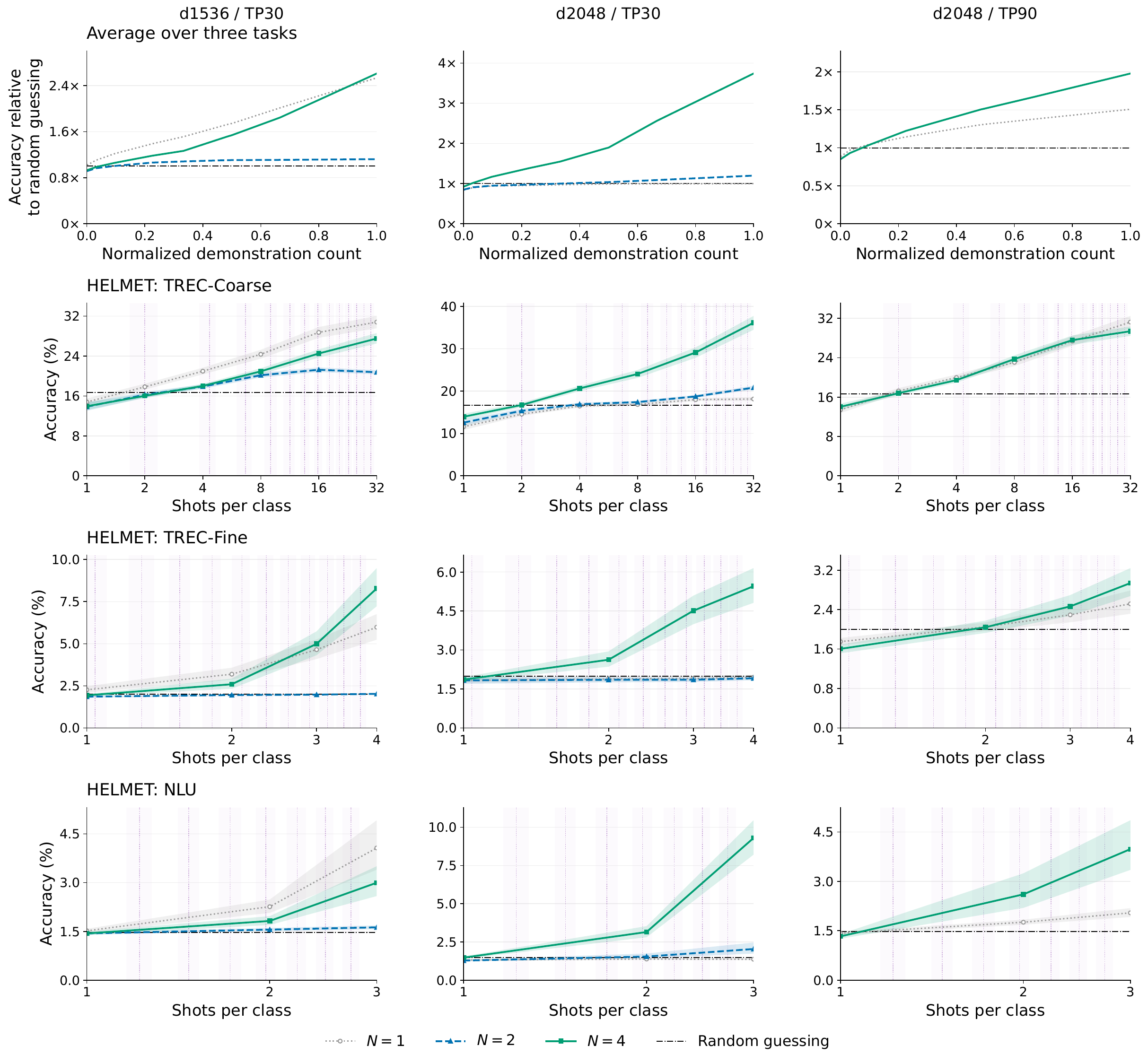}
    \caption{\textbf{In-context adaptation on three HELMET-ICL tasks.} The top row shows the average accuracy divided by the random baseline's accuracy over all tasks, where x-axis is normalized between 0 and 1.
    Vertical dotted lines show median sleep boundaries, and the shaded areas around them represent the 1st--99th percentile range.
    }
    \label{fig:helmet-full}
\end{figure*}

\section{In-context adaptation tasks}
\label{sec:icl_tasks}

\subsection{NLL eval}
\textbf{Repo2025}
We use the following repositories:
\begin{itemize}[nosep]
  \item OpenCode (\url{https://github.com/anomalyco/opencode})
  \item Gemini CLI (\url{https://github.com/google-gemini/gemini-cli})
  \item Open Lovable (\url{https://github.com/firecrawl/open-lovable})
  \item Microsoft Edit (\url{https://github.com/microsoft/edit})
  \item FlashMLA (\url{https://github.com/deepseek-ai/FlashMLA})
  \item OpenPipe ART (\url{https://github.com/OpenPipe/ART})
  \item tweakcn (\url{https://github.com/jnsahaj/tweakcn})
  \item DeepEP (\url{https://github.com/deepseek-ai/DeepEP})
  \item smallpond (\url{https://github.com/deepseek-ai/smallpond})
  \item DuckLake (\url{https://github.com/duckdb/ducklake})
  \item LangExtract (\url{https://github.com/google/langextract})
  \item ADK Python (\url{https://github.com/google/adk-python})
  \item DeepGEMM (\url{https://github.com/deepseek-ai/DeepGEMM})
\end{itemize}
. In total, these are 41,238,528 tokens.
We pack these to suitable lengths while ensuring each sequence is from one repository.

\textbf{arXiv2025}
We use \url{huggingface.co/datasets/SlowGuess/Arxiv_2025_OCR}, a collection of arXiv papers released in 2025 extracted as a markdown format.
We additionally remove Markdown image links. 
We keep every paper with at least 32,768
tokens, take its first 32,768 tokens, and split into eight independent 4k rows.
These are 42,860,544 tokens in total.

\subsection{Algorithmic word tasks}
Here, we provide three-shot examples of each word task.
Words used here are from the 10,000 most common English words~\footnote{\url{github.com/first20hours/google-10000-english/blob/d0736d492489198e4f9d650c7ab4143bc14c1e9e/google-10000-english.txt}}
. We further filtered to words that are one token both with and without a leading space.

\subsubsection{Selective copying}

\begin{lstlisting}[basicstyle=\ttfamily\small, breaklines=true, frame=single]
Input: updated echo ships [chair] spread sk bridge infrared
Output: chair
Input: rc gc spring [account] thin exists psi coat
Output: account
Input: pk af focused tuning checked shop word [birds]
Output: birds
Input: math though star sharing [egg] lab when rotation
Output:
\end{lstlisting}
Expected answer is \texttt{egg}. This example uses \texttt{[chair]} to mark the target word \texttt{chair}.
We explored seven total formatting methods:

\noindent\textbf{\# word}
\begin{lstlisting}
Input: updated echo ships # chair spread sk bridge infrared
Output: chair
\end{lstlisting}

\noindent\textbf{Asterisk: \texttt{* word}}
\begin{lstlisting}
Input: updated echo ships * chair spread sk bridge infrared
Output: chair
\end{lstlisting}

\noindent\textbf{[word]}
\begin{lstlisting}
Input: updated echo ships [chair] spread sk bridge infrared
Output: chair
\end{lstlisting}

\noindent\textbf{(word)}
\begin{lstlisting}
Input: updated echo ships (chair) spread sk bridge infrared
Output: chair
\end{lstlisting}

\noindent\textbf{keep / ignore}
\begin{lstlisting}
Input: ignore updated; ignore echo; ignore ships; keep chair; ignore spread; ignore sk; ignore bridge; ignore infrared
Output: chair
\end{lstlisting}

\noindent\textbf{select / skip}
\begin{lstlisting}
Input: skip updated; skip echo; skip ships; select chair; skip spread; skip sk; skip bridge; skip infrared
Output: chair
\end{lstlisting}

\noindent\textbf{yes / no}
\begin{lstlisting}
Input: no updated; no echo; no ships; yes chair; no spread; no sk; no bridge; no infrared
Output: chair
\end{lstlisting}

\begin{figure}
    \centering
    \includegraphics[width=1\linewidth]{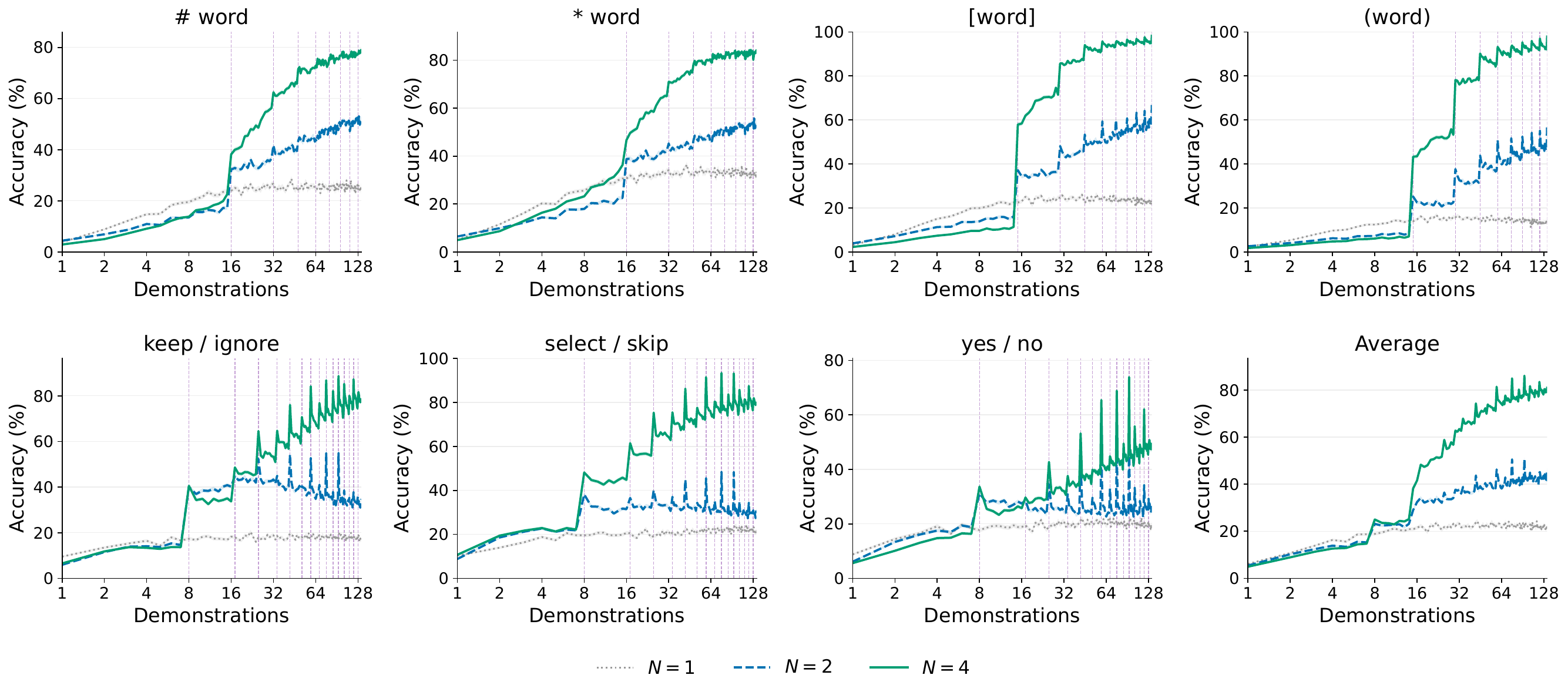}
    \caption{Selective recall accuracy with respect to the number of demonstrations depending on the formatting choice. 1.2B $N=1,2,4$ models trained on 32B tokens are used.}
    \label{fig:selective_copying_formatting}
\end{figure}

As shown in \Cref{fig:selective_copying_formatting}, the subtle choice of formatting leads to a large variation in performance, which further motivated us to focus on NLL evaluation as a more reliable signal.

\subsubsection{Reversal}

\begin{lstlisting}[basicstyle=\ttfamily\small, breaklines=true, frame=single]
Input: g r o w n
Output: n w o r g
Input: w h e r e
Output: e r e h w
Input: r e p a i r
Output: r i a p e r
Input: m e d
Output:
\end{lstlisting}
Expected answer is \texttt{d e m}.

\subsubsection{Latest assignment}

\begin{lstlisting}[basicstyle=\ttfamily\small, breaklines=true, frame=single]
Context:
gender is watching. scores is experienced. scores is confirmed. super is solved.
Input: scores
Output: confirmed
Context:
balls is asp. balls is oven. cr is icon. cm is units.
Input: balls
Output: oven
Context:
bound is rep. mess is dollar. bound is crime. suggest is other.
Input: bound
Output: crime
Context:
align is pleasant. align is local. pub is gt. should is city.
Input: align
Output:
\end{lstlisting}
Expected answer is \texttt{local}. 
This task tests whether the model can keep tracking of state overriding.

\subsubsection{Symbol removal}

\begin{figure}
    \centering
    \includegraphics[width=0.75\linewidth]{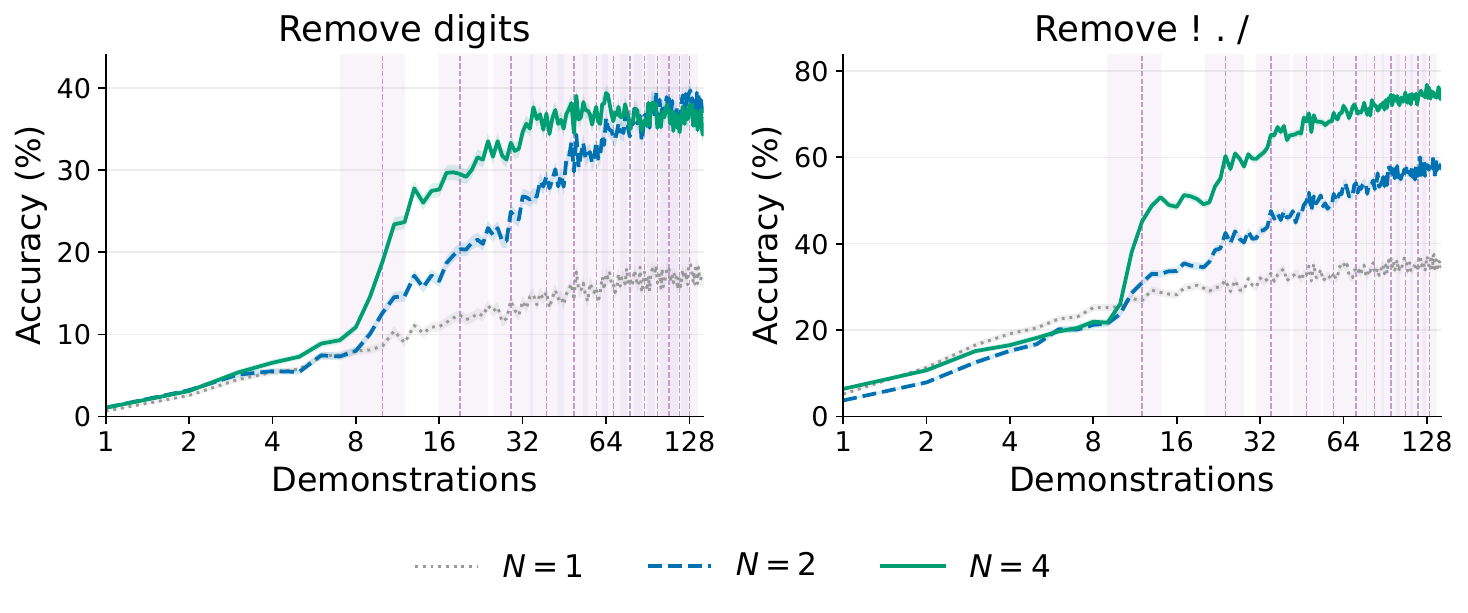}
    \caption{Symbol removal accuracy with respect to the number of demonstrations depending on the formatting choice. 1.2B N = 1, 2, 4
models trained on 32B tokens are used. Vertical dotted lines show median sleep boundaries, and the shaded areas around them represent the 1st--99th percentile range.}
    \label{fig:symbol_removal_formatting}
\end{figure}

\begin{lstlisting}[basicstyle=\ttfamily\small, breaklines=true, frame=single]
Input: p 4 l 4 a 4 n 5 e 1 s
Output: p l a n e s
Input: p 8 e 9 r 5 c 6 e 2 p 6 t 9 i 0 o 4 n
Output: p e r c e p t i o n
Input: p 7 o 0 p 1 u 6 l 2 a 2 t 8 i 2 o 2 n
Output: p o p u l a t i o n
Input: u 4 n 2 d 4 e 5 r 4 s 9 t 5 a 1 n 1 d
Output:
\end{lstlisting}
Expected answer is \texttt{u n d e r s t a n d}. 
Unlike \citet{brown2020language}, we insert spacing between each character, because the goal here is to test whether a model can recognize a symbol manipulating rule and execute it correctly, and we do not want to couple this with the effects of tokenizers and learned word embeddings.
Instead of inserting random digits, we can also insert special characters such as \texttt{.}, \texttt{!}, and \texttt{/} following \citet{brown2020language}.
As shown in in \Cref{fig:symbol_removal_formatting}, this choice leads to a large variance in accuracy, again motivating us to rely more on NLL evaluations.

\subsubsection{Cyclic rotation}

\begin{lstlisting}[basicstyle=\ttfamily\small, breaklines=true, frame=single]
Input: a p t
Output: a p t
Input: r s y e a
Output: y e a r s
Input: f a c e s u r
Output: s u r f a c e
Input: l e s c y c
Output:
\end{lstlisting}
Expected answer is \texttt{c y c l e s}. Every demonstration in this sequence rotates right by three letters.
Unlike \citet{brown2020language}, we insert spacing between each character.
Each sequence randomly samples the shift and shares it across demonstrations so the model can infer the shift from demonstrations.

\subsection{HELMET-ICL}
Here, we take three tasks from the widely used \citep{abdin2024phi,team2025kimi,zeng2026glm} ICL subset of HELMET dataset~\citep{yen2025helmet}: TREC-coarse, TREC-fine~\citep{li2002learning}, and NLU~\citep{liu2021benchmarking}.
These are text classification tasks with 6, 50, 68 labels respectively for TREC-coarse, TREC-fine, and NLU.
An example demonstration of TREC-coarse:

\begin{lstlisting}[basicstyle=\ttfamily\small, breaklines=true, frame=single]
What are some tips for building a fire in a fireplace ?
label: 4

How many Canadians emmigrate each year ?
label: 3

What do the letters CE stand for on so many products , particularly electrical , purchased now ?
label: 5

Why do movie and TV stars get paid so much ?
label: 2

What building are British monarchs crowned in ?
label: 1

Name the creator of `` The Muppets '' .
label: 0

How far is it from Denver to Aspen ?
label: 3
\end{lstlisting}
Demonstrations like this are presented to the model multiple times so the model can infer the classification rule.

\subsection{SST-2 nonsemantic}

\begin{lstlisting}[basicstyle=\ttfamily\small, breaklines=true, frame=single]
Input: seagal ran out of movies years ago , and this is just the proof .
Output: foo
Input: like the rugrats movies , the wild thornberrys movie does n't offer much more than the series , but its emphasis on caring for animals and respecting other cultures is particularly welcome .
Output: bar
Input: this is cruel , misanthropic stuff with only weak claims to surrealism and black comedy .
Output:
\end{lstlisting}
Expected answer is \texttt{ foo}. Here, the model has to infer that \texttt{foo} means negative, and \texttt{bar} means positive.
This task is introduced in \citet{wei2023larger} to decouple the model's in-context adaptation ability from learned semantic prior.

\end{document}